\documentclass[runningheads]{llncs}
\usepackage[T1]{fontenc}
\usepackage{graphicx}
\usepackage{svg}
\usepackage{xcolor}
\usepackage{amsmath}
\usepackage{amssymb}
\usepackage{multicol}
\usepackage{multirow}
\usepackage{soul}
\usepackage{booktabs}   
\usepackage{hyperref}
\usepackage{color}

\usepackage[capitalize]{cleveref}
\newcommand{\repeatthanks}{\textsuperscript{\thefootnote}}
\begin{document}
\title{A Framework for Low-Effort Training Data Generation for Urban Semantic Segmentation} 
%
\titlerunning{Low-Effort Training Data Generation for Urban Semantic Segmentation}
%
\author{Damjan Kal\v{s}an\thanks{Equal contribution}\inst{1} \and
Denis Zavadski\repeatthanks\inst{1,4} \and
Tim K\"uchler\inst{1} \and
Haebom Lee\inst{2} \and \\
Stefan Roth\inst{3,4,5} \and
Carsten Rother\inst{1,4}}
\authorrunning{D. Kal\v{s}an et al.}
%
\institute{Computer Vision and Learning Lab, IWR, Heidelberg University, Heidelberg, Germany \\
\email{\{damjan.kalsan, denis.zavadski, tim.kuechler, carsten.rother\}@iwr.uni-heidelberg.de} \and
AIMMO, Seongnam, Republic of Korea\\
\email{haebom.lee@gmail.com} \and
Department of Computer Science, TU Darmstadt, Darmstadt, Germany \\
\email{stefan.roth@visinf.tu-darmstadt.de} \and
Zuse School ELIZA, Darmstadt, Germany \and
Hessian Center for AI (hessian.AI), Darmstadt, Germany}

\maketitle              
\begin{abstract}
Synthetic datasets are widely used for training urban scene recognition models, but even highly realistic renderings show a noticeable gap to real imagery. This gap is particularly pronounced when adapting to a specific target domain, such as Cityscapes, where differences in architecture, vegetation, object appearance, and camera characteristics limit downstream performance. Closing this gap with more detailed 3D modelling would require expensive asset and scene design, defeating the purpose of low-cost labelled data.
To address this, we present a new framework that adapts an off-the-shelf diffusion model to a target domain using only imperfect pseudo-labels. Once trained, it generates high-fidelity, target-aligned images from semantic maps of any synthetic dataset, including low-effort sources created in hours rather than months. The method filters suboptimal generations, rectifies image-label misalignments, and standardises semantics across datasets, transforming weak synthetic data into competitive real-domain training sets.
Experiments on five synthetic datasets and two real target datasets show segmentation gains of up to +8.0\%pt. mIoU over state-of-the-art translation methods, making rapidly constructed synthetic datasets as effective as high-effort, time-intensive synthetic datasets requiring extensive manual design. This work highlights a valuable  collaborative paradigm where fast semantic prototyping, combined with generative models, enables scalable, high-quality training data creation for urban scene understanding.

\end{abstract}

\begin{figure*}[t]
  \centering
  \includegraphics[width=\linewidth]{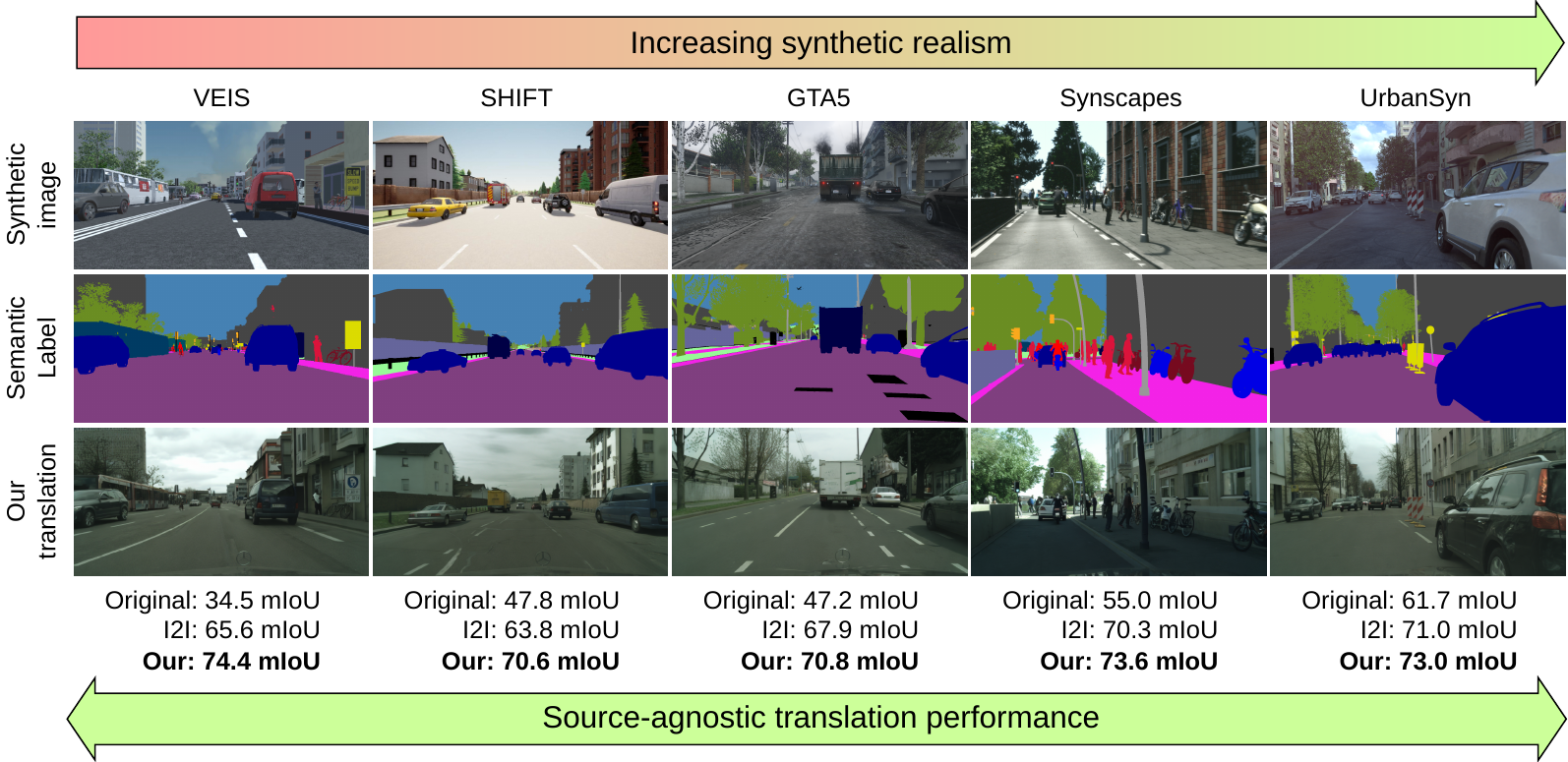}
  \caption{Given a synthetic source dataset \emph{(top row)} with the corresponding semantic labels \emph{(second row)}, we generate images \emph{(third row)} that adhere to the semantic map and lie in a particular target distribution, here Cityscapes~\cite{Cordts2016Cityscapes}. Below the images, we report the performance when training a downstream semantic segmentation system~\cite{NEURIPS2021_64f1f27b} on: \emph{i)} the synthetic source images (Original), \emph{ii)} using only generated target images from the best performing competitor method~\cite{zhu2017unpaired}, and \emph{iii)} only our generated images. Being source-agnostic, our approach performs equally well for all synthetic source datasets, regardless of their visual realism, thereby reducing the need for extensive 3D modelling effort and making low-effort datasets like VEIS~\cite{Saleh_2018_ECCV} a viable alternative.
  It outperforms the synthetic source data and all tested I2I methods by at least +2.0\%pt.\ mIoU.
  }
\label{fig:teaser}
\end{figure*}

\section{Introduction}
\label{sec:introduction}

Labelled training data is essential for semantic segmentation, yet collecting large-scale real-world annotations for urban driving scenes is costly and slow. Synthetic datasets offer a scalable alternative, but even the most photorealistic graphics leave a noticeable gap to real imagery, limiting downstream performance. Closing this gap by crafting high-quality 3D assets and detailed scenes, which are then rendered into 2D images, demands significant manual effort and deep expertise in computer graphics software, undermining the promise of cheap and scalable training data.

Recently, advances in diffusion models (DMs) such as Stable Diffusion~\cite{rombach2022high} and Flux~\cite{flux2024} have made high-quality, controllable image generation accessible to a wide audience. This development raises two important questions: can DMs replace the costly process of modelling photorealism in synthetic data? And, more broadly, should we foster a collaboration between the 3D modelling and generative modelling communities, where synthetic creators focus on rapidly producing diverse scene layouts with realistic geometry but simple appearance (i.e. low-effort synthetic data), while DMs translate these layouts into realistic data within a custom target domain at scale?

We address these questions with a diffusion-based framework that adapts to a target domain using only unlabelled real images and derived pseudo-labels. Once trained, it can translate semantic label maps from any source, including low-effort datasets or even manually composed layouts, into high-quality training data. Hence, our framework is source-agnostic.
We evaluate this by translating five synthetic datasets of varying visual realism to two real target domains.
Our experiments show that the proposed approach outperforms leading image-to-image (I2I) translation techniques in both visual quality, measured by CMMD~\cite{jayasumana2024rethinking}, and downstream semantic segmentation performance when trained exclusively on translated images. Importantly, when translating low-effort synthetic datasets such as VEIS~\cite{Saleh_2018_ECCV}, created in a single day, our generated images improve segmentation performance by up to +8.0\%pt. mIoU, matching or surpassing results obtained with costly, photorealistic synthetic datasets like UrbanSyn (see \cref{fig:teaser}).
Our approach also exceeds the performance of Synscapes~\cite{wrenninge2018synscapes}, which was explicitly designed to mimic the target-domain distribution of Cityscapes~\cite{Cordts2016Cityscapes}.
These results show that low-effort synthetic data, when translated with modern generative methods, can serve as high-quality real-domain training data.
This reduces the need for deep expertise in 3D modelling or time-intensive rendering and motivates a shift of focus towards modelling realistic geometry rather than visual appearance when creating synthetic datasets.

A concurrent line of work in unsupervised domain adaptation (UDA) tackles synthetic-to-real transfer differently. Given a labelled synthetic dataset, these methods directly train a semantic segmentation model on an unlabelled real dataset using iterative training on high-confidence pseudo-labels. Although UDA methods achieve strong results, they function as black boxes: they usually adapt a recogniser without ever producing intermediate target-domain images. This limits transparency, interpretability, and reusability of the resulting models. In contrast, our approach explicitly generates target-aligned images, which provides several practical advantages: \textit{(i) Transparency and auditability:} Generated images can be manually inspected or automatically checked for quality, which is critical for safety-sensitive vision applications and regulatory approval (e.g. as required in the European Union). \textit{(ii) Independent progress in data generation and recognition:} Our approach decouples data generation from downstream models, allowing separate improvements and broader reuse of datasets across different models, as well as perception tasks, such as detection or tracking. \textit{(iii) Rapid creation of rare or safety-critical scenarios:} New scenes can be synthesised from drawn or collaged semantic maps without constructing detailed 3D scenes, also reducing one major bottleneck of traditional synthetic data production.

In summary, we explore the potential collaboration between the 3D modelling and generative modelling communities to enable scalable creation of high-quality training datasets with reduced manual effort. Our contributions are:
\begin{itemize}
    \item We present a simple framework that adapts an off-the-shelf diffusion model to a specific target domain using unlabelled target images, imperfect pseudo-labels, and regularisation techniques. The framework can generate images from any synthetic dataset or manually composed semantic layout, and employs an object-centric filtering strategy to discard suboptimal generations.
    \item We conduct a large-scale evaluation analysis by translating five synthetic datasets to two real-world target domains and training two downstream semantic segmentation models exclusively on the generated data, resulting in over one hundred trained models in total.
    \item Data generated with our framework surpasses all competing image-to-image translation methods by up to +8\%pt. mIoU and achieves performance exceeding that of laboriously crafted photorealistic synthetic datasets. We release our generated data for others to use at \url{vislearn.github.io/FLEDGE/}.
\end{itemize}

\section{Related Work}
\label{sec:related_work}
\textbf{Image-to-image (I2I)} translation aims to map images from a source domain to a target domain while preserving semantic structure. While paired approaches exist~\cite{park2019semantic,wang2018high}, paired datasets are rarely available for synthetic-to-real transfer. Most methods therefore adopt unpaired strategies \cite{peng2023diffusionTranslation}, often based on adversarial training~\cite{cai2024rethinking,fu2019geometry,hoffman2018cycada,parmar2024one,richter2022enhancing,sankaranarayanan2018learning,zhu2017unpaired}, sometimes combined with cycle-consistency~\cite{zhu2017unpaired}, contrastive learning~\cite{park2020contrastive}, or content–style disentanglement~\cite{huang2018multimodal,lee2018diverse}.  
Despite their success in visual domain transfer, these approaches have two main limitations for downstream recognition tasks: (i) the translated images often lack realism, exhibiting artifacts or texture mismatches that reduce their utility for training segmentation models~\cite{richter2022enhancing}; and (ii) their performance strongly depends on the visual quality of the source images, leading to large drops when using low-effort synthetic data such as VEIS~\cite{Saleh_2018_ECCV}.  
Our method addresses these issues by adapting an off-the-shelf diffusion model without requiring source RGB images, using a source-agnostic training scheme. This allows the generation of high-quality, target-domain images from a variety of source distributions, including low-effort semantic layouts, and to generate entirely new scenes without paired images or 3D modelling. The resulting high-fidelity images can reliably serve as training data for downstream tasks such as urban semantic segmentation.

\textbf{Unsupervised Domain Adaptation (UDA)} seeks to adapt a model trained on a labelled source domain to perform well on an unlabelled target domain with a different data distribution.
Modern state-of-the-art UDA approaches~\cite{Hoyer_2022_CVPR,hoyer2022hrda,hoyer2023mic} typically operate as self-training pipelines, directly updating the recognition model without producing intermediate target-domain images. This limits transparency, prevents manual inspection of the adaptation process, and ties the adaptation results to a specific model and task.  
Some approaches, such as ControlUDA~\cite{shen2025wControlUDA}, attempt to generate target-domain images by conditioning on auxiliary cues like edge maps from source RGB images. However, these techniques remain dependent on high-quality source imagery and cannot synthesise new scenes from arbitrary or manually drawn semantic layouts.  
Our approach differs fundamentally by generating explicit target-domain images directly from semantic maps, without source RGB data. This enables visual inspection of the generated dataset, reuse across different recognition tasks, and on-demand creation of novel or rare scenarios --- capabilities that self-training UDA pipelines do not provide.

Recent work has explored \textbf{controlled image generation} with spatial signals such as semantic maps, using methods like concatenation~\cite{rombach2022high}, external control modules~\cite{mou2024t2i,zavadski2024controlnet,zhang2023adding}, classifier guidance~\cite{dhariwal2021diffusion}, or semantic infusion~\cite{jia2024dginstyle,kupyn2024dataset}. These techniques successfully enforce semantic consistency in generated images but generally fall short in two categories: (i) they do not consider a specific target domain, producing generic synthetic images for domain generalisation~\cite{jia2024dginstyle}, or (ii) they rely on real-domain inputs, as in ControlUDA~\cite{shen2025wControlUDA}, limiting applicability to synthetic-to-real transfer.  
Our approach instead fine-tunes an off-the-shelf diffusion model to a specific target domain using pseudo-labels, allowing high-quality target-aligned image synthesis from any synthetic semantic dataset. This makes it possible to translate low-effort synthetic datasets like VEIS into training data that matches or surpasses the effectiveness of expensive, laboriously engineered datasets such as UrbanSyn.

\section{Method}
\label{subsec:method}
We aim to translate synthetic semantic layouts $s_S$ from arbitrary source datasets into realistic images $x_\mathcal{T}$ that align with a specific real-world target domain, providing improved training data for semantic segmentation. We achieve this by adapting a pre-trained diffusion model through a source-agnostic fine-tuning pipeline, requiring only unlabelled target images and their estimated pseudo-labels. The framework consists of three parts: a) a two-stage fine-tuning strategy (Section~\ref{subsec:method-control_and_labeler}), b) regularisation techniques to improve robustness and source-agnostic generalisation (Section~\ref{subsec:method-regularisation}), and c) a large-scale data generation process with an object-centric selection mechanism (Section~\ref{subsec:method-filtering}).

\begin{figure*}[t]
  \centering
  \includegraphics[width=\linewidth]{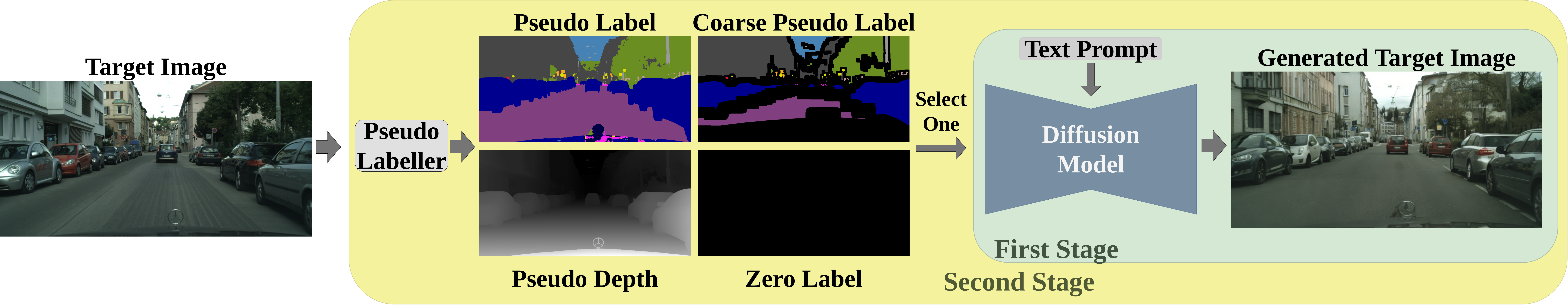}
  \caption{\emph{Overview of our two-stage training approach:} In the first stage, a pre-trained diffusion model is fine-tuned on unlabelled images from the (real) target domain. In the second stage, pseudo-labels predicted by a pre-trained method are used to further fine-tune the diffusion model for semantically-conditioned image synthesis. At test time, semantic maps from any source dataset can be used for the generation of target images.}
  \label{fig:training_pipeline}
\end{figure*}

\subsection{Fine-Tuning an Off-the-Shelf Diffusion Model for Image Translation}
\label{subsec:method-control_and_labeler}
We formulate synthetic-to-real translation as learning a conditional generative model $p_\theta(x|s)$  that approximates the target distribution $p_{\mathcal{T}}(x|s_S)$ without requiring source RGB images. The conditioning signal $s_S$ is a semantic segmentation map, readily available from synthetic data. During fine-tuning, we do not have ground-truth labels for the target images $x_\mathcal{T}\sim p_\mathcal{T}$; instead, similarly to ~\cite{shen2025wControlUDA}, we estimate pseudo-labels $\hat s_\mathcal{T} = f_L(x_\mathcal{T})$  using a pre-trained segmentation model $f_L$, providing the spatial structure for conditional generation. The training pipeline is illustrated in \cref{fig:training_pipeline}. 

Fine-tuning must solve two competing tasks: (i) learn the global visual style of the target domain, and (ii) align the generated image with the semantic map. Training both objectives jointly can cause trade-offs or unstable optimisation, where neither target style nor semantic fidelity is fully captured. We mitigate this with a two-stage training scheme.
\textit{Stage 1 (Target appearance adaptation):} We first align the diffusion model with the target domain distribution by minimising the standard noise prediction loss

\begingroup
\small    
\begin{align}
    \mathcal{L}_{\text{style}}(\theta)= \mathbb{E}_{x_t,\epsilon,t}\big[\lVert \epsilon - \epsilon_\theta(x_t,t)\rVert_2^2\big],
\end{align} 
\endgroup

using unlabelled target images $x_\mathcal{T}$ and automatically generated captions~\cite{JoyCaption} for text conditioning.
\textit{Stage 2 (Semantic conditioning):} Once the model captures target textures and scene statistics, we introduce pseudo-label conditioning $\hat{s}_\mathcal{T}$:

\begingroup
    \small
\begin{align}
\mathcal{L}_{\text{cond}}(\theta)= \mathbb{E}_{x_t,\hat{s}_\mathcal{T},\epsilon,t}\big[\lVert \epsilon - \epsilon_\theta(x_t,\hat{s}_\mathcal{T},t)\rVert_2^2\big].    
\end{align}
\endgroup
This sequential optimisation stabilises training and improves adherence to the semantic layout.
Spatial conditioning can be implemented either with auxiliary control networks~\cite{zavadski2024controlnet,zhang2023adding} or by concatenating semantic maps with the input, as in~\cite{rombach2022high}. We empirically find the latter approach to yield superior semantic fidelity (see~\cref{tab:ablation}).

\subsection{Regularisation Techniques}
\label{subsec:method-regularisation}
In the second training stage, we rely on semantic pseudo-labels estimated from the unlabelled target domain to provide semantic conditioning.
Pseudo-labels $\hat{s}_\mathcal{T}$ are imperfect: they may contain errors, missing objects, or dataset-specific structural biases. Training solely on detailed masks can make the model sensitive to noise and over-specialised to the style of one particular pseudo-labeller, hindering generalisation to unseen synthetic sources. To address this, we occasionally replace $\hat{s}_\mathcal{T}$ with a coarse map $\hat s_\mathcal{T}^\prime$ where uncertain boundaries are removed via erosion:
    \begingroup
        \small
    \begin{align}
        \hat s_\mathcal{T}^\prime = \text{Erode}(\hat{s}_\mathcal{T}, \lambda |\kappa|), \quad \lambda=0.15,
    \end{align}
    \endgroup
for each connected component $\kappa$ comprised from $|\kappa|$ pixels using a circular kernel with radius $\lambda|\kappa|$ proportional to its size.
This teaches the model to prioritise stable, visually supported spatial structures, improving robustness to label noise and source-agnostic generalisation across datasets.

To further encourage better spatial reasoning and prevent overfitting on limited target data, we occasionally replace the semantic conditioning with a pseudo-depth map $\hat{d} = f_D(x_\mathcal{T})$ with a probability of 20\%.
Depth maps encode geometric layout similar to semantic maps but in a different, continuous format, exposing the model to alternative structural cues. This acts as input-level regularisation, improving robustness to noisy masks and enabling generalisation to diverse semantic input styles.

In diffusion-based generation, classifier-free guidance (CFG)~\cite{ho2022classifier} improves sample quality by extrapolating the difference between predictions with and without conditioning. In its classical formulation, CFG enforces alignment with a text prompt. In our case, text control is less relevant, as image structure is guided by a semantic map and content is unconditionally learned by fine-tuning on the target domain. Still, we use CFG to improve adherence of our generation to the provided semantic map by utilising a zero conditioning in form of a black image (i.e., empty spatial condition) in 10\% of training steps. This strategy leads to stronger semantic consistency and yields more reliable synthetic-to-real translations, as confirmed experimentally.

\subsection{Automated Training Data Generation}
\label{subsec:method-filtering}
\sloppy
Conventional I2I approaches~\cite{park2020contrastive,parmar2024one,richter2022enhancing,zhu2017unpaired} produce deterministic translations with strict pixel-wise alignment between source and output images. While this enables direct reuse of original semantic labels, it limits scalability: only one translation per input is possible, thereby limiting dataset diversity.
Unlike deterministic I2I methods, our diffusion model can generate diverse samples $\{x_i\}_{i=1}^N \sim p_\theta(x|s_S)$ for a single semantic map. Since perfect pixel alignment cannot be guaranteed, we re-estimate pseudo-labels $\hat{s}_i=f_L(x_i)$ from the translated samples and rank candidates by our Mean Class-wise Object Consistency (MCOC).
For each connected component $\kappa_j$ in $s_S$ we compute the dominant number of pixels $\alpha_{c}(\kappa_j)$ predicted in $\hat s_i$ as class $c\in C$:
\begingroup
    \small
\begin{align}
    \alpha_{c}(\kappa_j)=\frac{|\{\text{pixel}\in \kappa_j\mid \hat{s}(\text{pixel})=c\}|}{|\kappa_j|},
\end{align}
\endgroup
The component is accepted if $\max_c \alpha_{c}(\kappa_j)\geq\tau$ $(\tau=0.7)$ meaning its predicted label is sufficiently reliable and dominated by a single class; otherwise it is rejected. With per-class acceptance scores
\begingroup
    \small
 \begin{align}
     A_c^{\text{comp}}=\frac{\text{\# accepted components for }c}{\text{\# total components for }c}
 \end{align}
\endgroup
we define the MCOC score for a sample as:
    \begingroup
        \small
        \begin{align}
            \text{MCOC}(x_i)=\frac{1}{|C|}\sum_{c\in C} A_c^{\text{comp}}
        \end{align}
    \endgroup
averaging over classes present in $s_S$ to avoid dominance by frequent ones (e.g., road, sky). We generate $N$ samples, select the top $k$ by MCOC, and pair them with their pseudo-labels for training. This allows us to improve both diversity and semantic reliability of the automatically constructed training data.

\section{Experiments}

\subsection{Experimental Setup}
\label{subsec:experimental_setup}

\textbf{Datasets.} 
We translate five labelled synthetic datasets of varying image quality (see \cref{fig:teaser}) to two unlabelled target domains, Cityscapes~\cite{Cordts2016Cityscapes} and ACDC~\cite{sakaridis2021acdc}. 
For synthetic datasets,
UrbanSyn (7539 images)~\cite{GOMEZ2025130038} and Synscapes (25000 images) \cite{wrenninge2018synscapes} exhibit high realism and best resemble our target datasets.
GTA5 (24966 images) \cite{Richter_2016_ECCV} is extracted from a popular video game with industry-grade graphics in US cities.
Lastly, SHIFT~\cite{sun2022shift} and VEIS~\cite{Saleh_2018_ECCV} are of simple synthetic quality.
Notably, VEIS exemplifies a low-effort dataset, having been created by a single person within one day.
To reduce the dataset size of SHIFT and VEIS, we select a subset of 3000 and 3018 images, respectively. For details on subset creation we refer to the supplement \ref{sup_subsec:subset_creation}.
The real target datasets, Cityscapes and ACDC, are captured in German and Swiss cities, respectively.
Cityscapes consists of four subsets: train (2975 images), validation (500 images), test (1525 images), and train-extra (20000 images). All images are captured in normal daytime conditions.
In contrast, ACDC contains 1600 training and 406 validation images, each equally split between four adverse conditions: fog, rain, night, and snow.

\textbf{Implementation details.}
We choose Stable Diffusion 2.1 \cite{rombach2022high} as the pre-trained generative model and fine-tune it as described in \cref{subsec:method-control_and_labeler,subsec:method-regularisation}. We use HRDA~\cite{hoyer2022hrda} as the pseudo-labeller $f_L$ for each target dataset. Following~\cite{GOMEZ2025130038}, we train HRDA on a combination of three labelled synthetic source datasets (GTA5, Synscapes, and UrbanSyn) and the unlabelled training set of the corresponding target dataset.
For fine-tuning our model to Cityscapes, pseudo-labels are computed on the train-extra set. The same target images are used to train the competing methods.
For ACDC as target domain, our Cityscapes model is further fine-tuned on the ACDC training set (combining all four conditions). The competing I2I methods are trained exclusively on the ACDC training set for each adverse condition and source dataset separately, as they are not designed to work on multiple conditions jointly.
We train the competitors~\cite{cai2024rethinking,huang2018multimodal,parmar2024one,zhu2017unpaired} using their official codebases and training settings until convergence.
For Photorealism Enhancement~\cite{richter2022enhancing}, we adapted their code to only use RGB images and depth maps, as other buffers are not available.
All competitors are trained as image-to-image translation methods, 
except EnCo \cite{cai2024rethinking}, which considers unpaired label-to-image.
We evaluate the visual quality of the generated images using the FID~\cite{heusel2017gans} and the increasingly popular CMMD~\cite{jayasumana2024rethinking} scores.
We train two downstream models, SegFormer~\cite{NEURIPS2021_64f1f27b} and DeepLabV3+~\cite{chen2018encoder}, and report the mean intersection over union (mIoU) on the target validation set.
To isolate the effect of data translation, we train the downstream models exclusively on the translated data.
All downstream training experiments are using pseudo-labels for the translated data. We observe that using the original synthetic labels instead of pseudo-labels consistently reduces performance across methods; corresponding results are provided in the supplement \ref{sup_subsec:additional_quantitative}.
For further details on training the downstream task, please refer to the supplement \ref{sup_subsec:downstream_details}.
For depth-map regularisation of our method, we use Depth~Anything~V2~\cite{yang2024depth}.

\subsection{Quantitative and Qualitative Comparison}
\label{subsec:quantitative_and_qualitative_comparison}

The results on the downstream semantic segmentation task are shown in \cref{tab:sota-comparison-downstream}. In contrast to I2I competitors, whose performance is correlated with synthetic data realism, our method performs roughly equally well over all source datasets, demonstrating its source-agnostic characteristic.
For Cityscapes, our method outperforms the strongest competitor in 9 out of 10 evaluated setups, with gains ranging between +2.0 and +8.0\%pt.\ mIoU for SegFormer, and up to +7.9\%pt.\ mIoU for DeepLabV3+.
Notably, even when translating from VEIS, a low-effort dataset constructed in just one day, our method achieves 74.4\%pt. mIoU and outperforms the strongest performing competitor~\cite{richter2022enhancing,zhu2017unpaired} by +3.4\%pt.\ mIoU, despite the latter using highly-realistic UrbanSyn as source.
This highlights the effectiveness of our proposed paradigm: generative methods combined with rapidly created synthetic scenes can outperform laborious design of visual realism.

\begin{table*}
    \caption{\emph{Comparison on semantic segmentation performance} (mIoU in \%, $\uparrow$) of our approach to five competing image translation methods, translating from five synthetic datasets to Cityscapes~\cite{Cordts2016Cityscapes} and ACDC~\cite{sakaridis2021acdc}. Methods marked with \dag~can generate multiple diverse images per synthetic sample. For $\rightarrow$ Cityscapes, we generate three images per sample for these methods. All other methods only generate one image per sample due to deterministic restrictions. For $\rightarrow$ ACDC, all methods generate a single image per sample. The best approach is highlighted in \textbf{bold}, the second best \underline{underlined}. For per-class results, please refer to the supplement \ref{sup_subsec:additional_quantitative}.}
    \label{tab:sota-comparison-downstream}
    \centering
    \footnotesize
    \resizebox{\textwidth}{!}{
    \begin{tabular}{@{}clccccccc@{}}
    \toprule
    Downstr. & Translation & \multicolumn{5}{c}{$\rightarrow$ Cityscapes (mIoU in \%, $\uparrow$)} & \multicolumn{2}{c}{$\rightarrow$ ACDC (mIoU in \%, $\uparrow$)} \\
    \cmidrule(lr){3-7} \cmidrule(lr){8-9}
    Model & Method & VEIS & SHIFT & GTA5 & Synscapes & UrbanSyn & VEIS & UrbanSyn \\
    \midrule
    \multirow{8}{*}{\rotatebox[origin=c]{90}{SegFormer}}
        & Original
            & 34.5 
            & 47.8 
            & 47.2 
            & 55.0 
            & 61.7 
            & 18.5 
            & 34.0 
            \\
        & CycleGAN \cite{zhu2017unpaired}
            & 65.6 
            & 63.8 
            & \underline{67.9} 
            & \underline{70.3} 
            & \underline{71.0} 
            & 44.7 
            & \underline{49.8} 
            \\
        & MUNIT\textsuperscript{\dag} \cite{huang2018multimodal}        
            & \underline{66.4} 
            & \underline{64.9} 
            & 65.8 
            & 67.3 
            & 70.9 
            & 46.4 
            & 49.0 
            \\
        & Ph. Enhanc. \cite{richter2022enhancing}  
            & 62.4 
            & 61.3 
            & 64.6 
            & 68.4 
            & \underline{71.0} 
            & \underline{48.5} 
            & 47.5 
            \\
        & I2I-Turbo \cite{parmar2024one}    
            & 60.0 
            & 61.5 
            & 63.4 
            & 64.4 
            & 69.6 
            & 43.0 
            & 48.1 
            \\
        & EnCo \cite{cai2024rethinking}        
            & 34.3 
            & 34.9 
            & 32.9 
            & 33.2 
            & 29.1 
            & 28.2 
            & 28.0 
            \\
        \cmidrule(lr){2-9}
        & Ours\textsuperscript{\dag}         
            & \textbf{74.4} 
            & \textbf{70.6} 
            & \textbf{70.8} 
            & \textbf{73.6} 
            & \textbf{73.0} 
            & \textbf{50.3} 
            & \textbf{50.4} 
            \\    
    \midrule
    \multirow{8}{*}{\rotatebox[origin=c]{90}{DeepLabV3+}}
        & Original
            & 19.0 
            & 44.2 
            & 31.6 
            & 45.3 
            & 47.8 
            & 10.5 
            & 14.4 
            \\
        & CycleGAN \cite{zhu2017unpaired}
            & \underline{57.8} 
            & \underline{55.3} 
            & \underline{52.4} 
            & 55.4 
            & 58.7 
            & 34.0 
            & 32.7 
            \\
        & MUNIT\textsuperscript{\dag} \cite{huang2018multimodal}        
            & 56.0 
            & 52.2 
            & 47.0 
            & 54.4 
            & \textbf{61.8} 
            & \textbf{36.4} 
            & \textbf{36.1} 
            \\
        & Ph. Enhanc. \cite{richter2022enhancing}  
            & 46.7 
            & 52.7 
            & 43.7 
            & \underline{56.2} 
            & \textbf{61.8} 
            & 30.7 
            & \textbf{36.1} 
            \\
        & I2I-Turbo \cite{parmar2024one}    
            & 50.1 
            & 51.1 
            & 48.1 
            & 53.9 
            & 60.0 
            & 29.4 
            & 33.1 
            \\
        & EnCo \cite{cai2024rethinking}        
            & 29.0 
            & 30.2 
            & 26.6 
            & 26.2 
            & 23.5 
            & 23.0 
            & 20.5 
            \\
        \cmidrule(lr){2-9}
        & Ours\textsuperscript{\dag}      
            & \textbf{62.3} 
            & \textbf{58.3} 
            & \textbf{55.8} 
            & \textbf{64.1} 
            & \underline{60.8} 
            & \underline{34.3} 
            & \underline{34.6} 
            \\    
    \bottomrule
    \end{tabular}
    }
\end{table*}

\begin{figure}[t]
  \centering
  \includegraphics[width=1.0\linewidth]{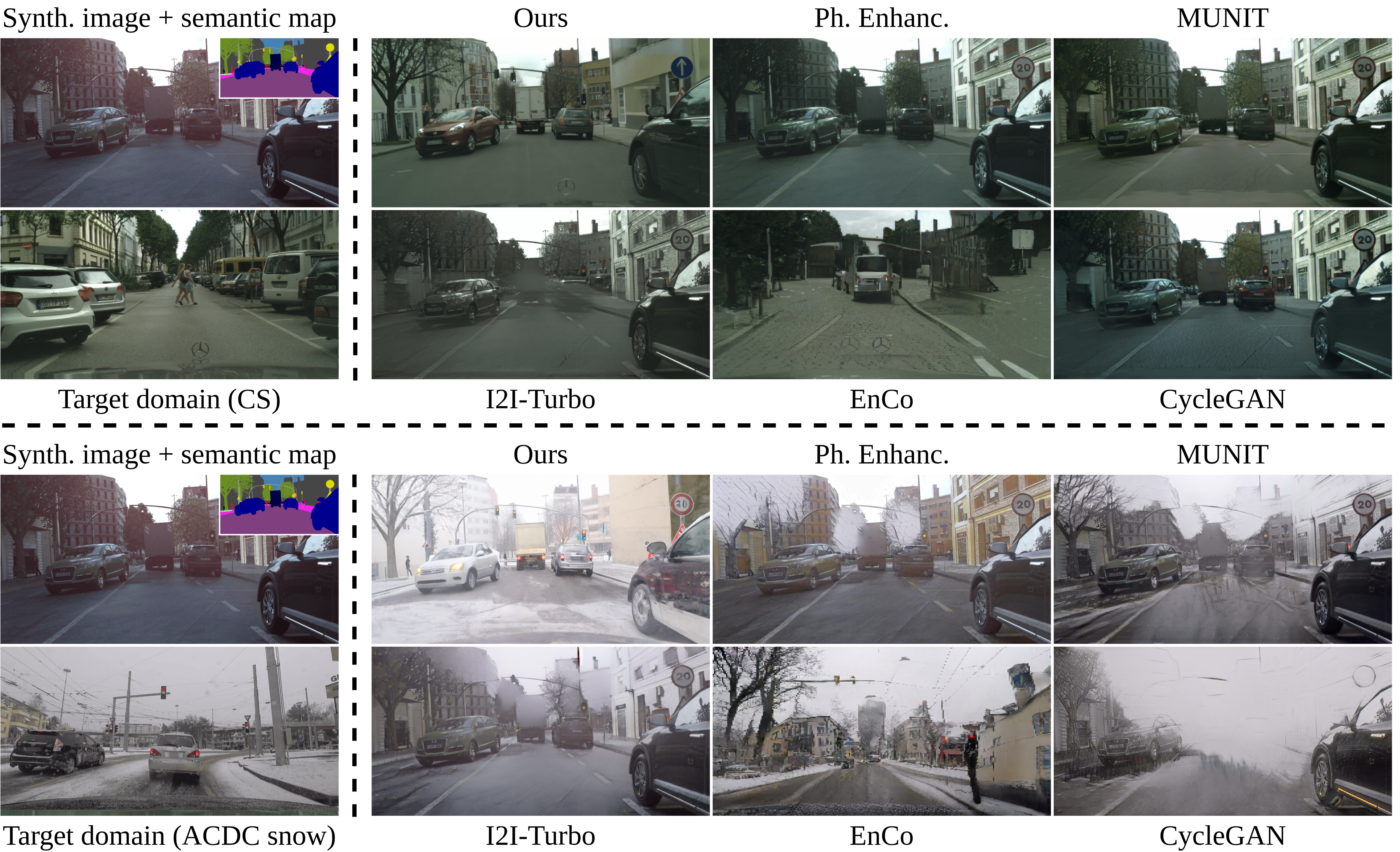}
  \caption{\emph{Comparison of images translated from UrbanSyn~\cite{GOMEZ2025130038} to Cityscapes~\cite{Cordts2016Cityscapes} \emph{(top)} and to ACDC snow \cite{sakaridis2021acdc} \emph{(bottom)}}. Our method features new objects and textures that closely align with the target datasets, while the competitors mostly transform only colours and show many artifacts for complex translations \emph{(bottom)}.}
  \label{fig:main_qualitative}
\end{figure}

On ACDC, our method performs favourably for SegFormer with gains up to +1.8\%pt. mIoU, but takes the second place for DeepLabV3+.
This brings up an interesting observation that practitioners should take caution when interpolating the performance of generative data from one downstream model to another.
Interestingly, EnCo learns to ignore the input source labels, as shown in \cref{fig:main_qualitative} in the bottom row, and hence performs poorly.
We conjecture this is due to the difference in camera perspectives between the source and target datasets, making it easy for the adversarial discriminator to tell apart real data from a translated image following the semantic layout.

We additionally compare our method to two diffusion-based methods, DGInStyle~\cite{jia2024dginstyle} and Instance Augmentation~\cite{kupyn2024dataset} and outperform them by a large margin (see supplement \ref{sup_sec:dginstyle_and_instaug}).
Moreover, including our translated data can even improve the performance of our pseudo-labeller method HRDA.
We achieve this by translating UrbanSyn to Cityscapes, adding the translated data (with labels) to the original three labelled datasets, and retraining the pseudo-labeller.
This increases the performance from 75.9\% to 76.5\% mIoU.

Visually, our method sets itself apart from the competitors by producing images with much higher realism, unprecedented creativity, and fewer artefacts (see \cref{fig:main_qualitative}).
Translating to Cityscapes, it closely captures the scene layout of the synthetic image and generates realistic variations of the scene, which could have easily come from the Cityscapes distribution.
In contrast, the competing methods make minimal changes to the original synthetic image, mainly only aligning the global colour distribution and making small textural changes.
For I2I-Turbo \cite{parmar2024one}, we consistently observe ``transparency'' artefacts, while EnCo does not follow the semantic layout.
The large gap in visual quality is also captured quantitatively in \cref{tab:sota-comparison-visual}, where our method leads on the CMMD metric by a large margin.
Notably, EnCo reaches one of the lowest FID scores while exhibiting the most visual inaccuracies and operating at the lowest resolution.
Note that FID has been criticised for various flaws~\cite{jayasumana2024rethinking,parmar2022aliased} which have since been addressed in the CMMD metric.
Thus, we cast doubt on method comparability using the FID score, but still present it for completeness.

\begin{table*}
    \caption{\emph{Comparison on image visual quality} of our approach to five competing image translation methods, translating from five synthetic datasets to Cityscapes~\cite{Cordts2016Cityscapes}. The best approach is highlighted in \textbf{bold}, the second best \underline{underlined}.}
    \label{tab:sota-comparison-visual}
    \centering
    \resizebox{\textwidth}{!}{
    \begin{tabular}{@{}lcccccccccc@{}}
    \toprule
      &  \multicolumn{2}{c}{VEIS} & \multicolumn{2}{c}{SHIFT} & \multicolumn{2}{c}{GTA5} & \multicolumn{2}{c}{Synscapes} & \multicolumn{2}{c@{}}{UrbanSyn}\\
      \cmidrule(lr){2-3}\cmidrule(lr){4-5}\cmidrule(lr){6-7}\cmidrule(lr){8-9}\cmidrule(lr){10-11}
     Method & CMMD$\,\downarrow$ & FID$\,\downarrow$ & CMMD$\,\downarrow$ & FID$\,\downarrow$ & CMMD$\,\downarrow$ & FID$\,\downarrow$ & CMMD$\,\downarrow$ & FID$\,\downarrow$ & CMMD$\,\downarrow$ & FID$\,\downarrow$\\
    \midrule
    Original                            
                                        & 4.517    
                                        & 128.0    
                                        & 4.996    
                                        & 287.5    
                                        & 5.182    
                                        & 79.2    
                                        & 2.407    
                                        & 41.1    
                                        & 3.290    
                                        & 50.7    
                                        \\
    CycleGAN \cite{zhu2017unpaired}
                                        & \underline{2.036}    
                                        & \underline{50.5}    
                                        & 2.327    
                                        & \underline{44.5}    
                                        & 2.313    
                                        & \underline{30.9}    
                                        & 1.582    
                                        & \underline{25.5}    
                                        & \underline{1.500}    
                                        & 26.4    
                                        \\
    MUNIT \cite{huang2018multimodal}                                                           
                                        & 2.919    
                                        & 53.1    
                                        & 3.346    
                                        & 53.1    
                                        & 3.252    
                                        & 38.4    
                                        & 1.395    
                                        & 31.6    
                                        & 1.650    
                                        & 30.8    
                                        \\
    Ph. Enhanc. \cite{richter2022enhancing}                                        
                                        & 3.747    
                                        & 94.0    
                                        & 3.669    
                                        & 75.7    
                                        & 3.413    
                                        & 46.9    
                                        & 1.564    
                                        & 37.2    
                                        & 1.777    
                                        & 34.4    
                                        \\
    I2I-Turbo \cite{parmar2024one}
                                        & 3.491    
                                        & 53.9    
                                        & 3.186    
                                        & 54.6    
                                        & 3.766    
                                        & 33.3    
                                        & \underline{1.279}    
                                        & 32.1    
                                        & 1.836    
                                        & 26.8    
                                        \\
    EnCo \cite{cai2024rethinking}       & 2.262    
                                        & 55.1    
                                        & \underline{1.759}    
                                        & \textbf{26.3}    
                                        &  \underline{1.623}   
                                        & \textbf{20.5}    
                                        & 1.884    
                                        & \textbf{21.3}    
                                        & 1.869    
                                        & \textbf{24.5}    
                                        \\
    \midrule
    Ours
                                        & \textbf{0.758}    
                                        & \textbf{42.3}    
                                        & \textbf{1.046}    
                                        & 49.2    
                                        & \textbf{0.933}    
                                        & 33.3    
                                        & \textbf{0.818}    
                                        & 31.5    
                                        & \textbf{0.659}    
                                        & \underline{24.7}    
                                        \\
    \bottomrule
    \end{tabular}
    }
\end{table*}

\begin{table*}
    \caption{\emph{Impact of object-centric sample ranking on semantic segmentation performance} (mIoU in \%, $\uparrow$) on Cityscapes~\cite{Cordts2016Cityscapes} using SegFormer~\cite{NEURIPS2021_64f1f27b}. Out of 10 generated samples for each semantic map, $k$ are chosen randomly or according to the proposed MCOC score.}
    \label{tab:mcoc}
    \footnotesize
    \centering
    \resizebox{\textwidth}{!}{
    \setlength{\tabcolsep}{1mm}
    \begin{tabular}{@{}lcccccccccc@{}}
    \toprule
     & \multicolumn{5}{c}{Top $k$ of 10 according to MCOC score} & \multicolumn{5}{c@{}}{Random $k$ of 10} \\
    \cmidrule(lr){2-6} \cmidrule(lr){7-11}
     $k$ & VEIS & SHIFT & GTA5 & Synscapes & UrbanSyn & VEIS & SHIFT & GTA5 & Synscapes & UrbanSyn \\
    \midrule
     1 & 
        72.8 &
        69.0 &
        70.1 &
        73.8 &
        72.9 &
        73.2 &
        67.6 &
        70.2 &
        73.3 &
        72.4
        \\
     3 & 
        74.4 &
        70.6 &
        70.8 &
        73.6 &
        73.0 &
        74.2 &
        70.7 &
        70.6 &
        73.2 &
        73.0
        \\
    \bottomrule
    \end{tabular}
    }
\end{table*}

Translating to ACDC is less satisfactory for all methods (see bottom row in \cref{fig:main_qualitative}).
There is still a large gap in realism, and artefacts can be observed for all methods.
This likely stems from the increased visual difficulty of the ACDC dataset, as well as significantly fewer unlabelled training images. Indeed, this is an open research direction deserving more attention from the community.

\begin{figure}[t]
  \centering
  \includegraphics[width=\linewidth]{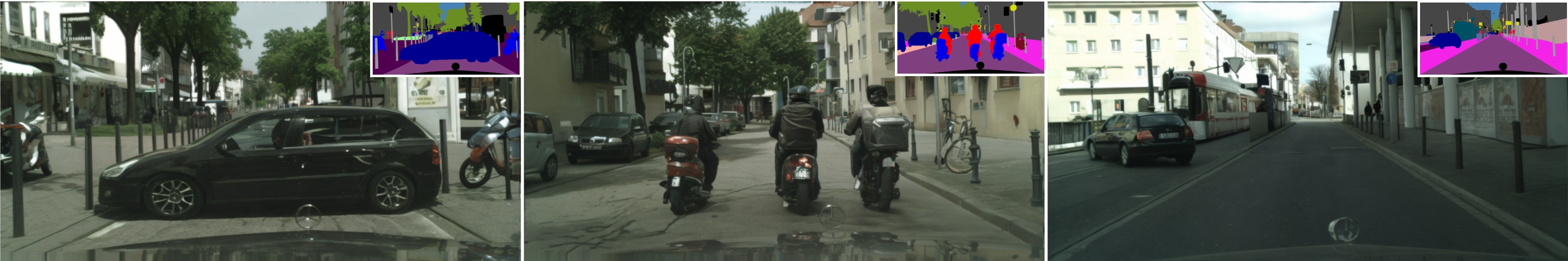}
  \caption{\emph{Generation of images of edge case scenarios from manually created semantic maps} with our approach, enabling the quantitative and qualitative analysis of such scenarios and use in training safety-critical systems. Note that these images are not part of Cityscapes~\cite{Cordts2016Cityscapes}.}
  \label{fig:special_scenarios}
\end{figure}

With the highly realistic image generation, as well as the faithfulness to the semantic label guidance, our method opens avenues for creating unseen traffic scenarios in the target domain.
We demonstrate this in \cref{fig:special_scenarios}, where we manually create three scenarios that never occurred in the original Cityscapes dataset.
Such generated data can be used to test the performance of existing recognisers in rare scenarios~\cite{loiseau2024reliability}, which are critical but difficult to gather in the real world.

\subsection{Ablation Study}
\label{subsec:ablation}
We validate our methodological decisions on the downstream performance of SegFormer, using VEIS and UrbanSyn and translating them to Cityscapes (see \cref{tab:ablation}).
In each experiment, only one component is replaced and compared to the ``Full'' method.
Concatenating pseudo-labels in the second fine-tuning step performs favourably compared to training a separate control model.
We verify this by replacing concatenation with ControlNet-XS~\cite{zavadski2024controlnet} and observe a significant drop in performance on both datasets (third column).
Similarly, omitting the first fine-tuning step and instead training the off-the-shelf model in one step directly on the pseudo-labels (fourth column) leads to suboptimal results.
This demonstrates that decoupling the learning of visual appearance and spatial control is a crucial design choice.
One of the largest performance drops is observed when we supervise SegFormer training using the original semantic labels instead of pseudo-labels (fifth column).
After careful inspection, we found that while our model faithfully follows the semantic conditioning most of the time, it sometimes generates semantically incorrect classes.
For example, it may generate a train instead of a bus, since these concepts and their masks are very similar (see supplement \ref{sup_subsec:additional_qualitative}).
Additionally, there are semantic inconsistencies between datasets (e.g., pickup trucks are labelled as truck in GTA5 and as car in Synscapes).
This can lead to semantic confusion in the downstream model, reducing the segmentation performance.
Both the black-image control and MCOC ranking (\cref{tab:mcoc}) generally improve segmentation performance across datasets, with no substantial drawbacks, and are therefore retained in our final method.

\begin{table}
    \caption{\emph{Ablation results reported on semantic segmentation performance} (mIoU in \%, $\uparrow$) on Cityscapes~\cite{Cordts2016Cityscapes} using SegFormer~\cite{NEURIPS2021_64f1f27b}. ``Full'' represents our proposed method, ``with CNXS'' denotes replacing concatenation in the second stage with ControlNet-XS~\cite{zavadski2024controlnet}, and ``w/o CFG'' denotes not using a black image as negative guidance.}
    \label{tab:ablation}
    \centering
    \footnotesize
    \setlength{\tabcolsep}{1mm}
    \begin{tabular}{lccccc}
    \toprule
    Source & Full & with CNXS & w/o first stage & with synthetic labels & w/o CFG \\
    \midrule
    VEIS 
        & 72.8  
        & 70.6  
        & 71.9  
        & 59.2  
        & 73.0  
        \\
    UrbanSyn
        & 72.9  
        & 70.5  
        & 71.0  
        & 66.4  
        & 71.9  
        \\
    \bottomrule
    \end{tabular}
\end{table}

\section{Conclusion}
\label{sec:conclusion}
We presented a source-agnostic framework that adapts an off-the-shelf diffusion model to transform low-effort semantic layouts into high-quality, target-domain training images for urban semantic segmentation. Using imperfect pseudo-labels and a two-stage fine-tuning strategy, our method generates realistic, spatially faithful images without paired supervision or reliance on source RGB inputs.
Experiments on five synthetic sources and two real targets show substantial gains over state-of-the-art image-to-image translation methods, achieving up to +8.0\%pt. mIoU improvement. Crucially, it demonstrates that synthetic datasets created in a single day can, when translated with our framework, rival laboriously engineered photorealistic datasets, substantially reducing the cost and expertise required for dataset creation.
This work highlights a promising paradigm, where fast semantic scene prototyping and generative diffusion models together enable scalable, high-quality data generation. We hope it inspires future works that further strengthen this collaborative paradigm, thus democratising access to large, high-quality datasets for a broad range of vision tasks.

\subsubsection{Acknowledgements.}
Denis Zavadski is supported by the Konrad Zuse School of Excellence in Learning and Intelligent Systems (\href{https://eliza.school/}{ELIZA}) through the DAAD programme Konrad Zuse Schools of Excellence in Artificial Intelligence, sponsored by the Federal Ministry of Education and Research. The authors gratefully acknowledge the support by the Ministry of Science, Research and the Arts Baden-Württemberg (MWK) through bwHPC, SDS@hd and the German Research Foundation (DFG) through the grants INST 35/1597-1 FUGG and INST 35/1503-1 FUGG.

%
%
%
\bibliographystyle{splncs04}
\bibliography{main}


\clearpage
\appendix
\section*{Supplementary Material}

\section{Implementation and Technical Details}

This section includes additional information to \cref{subsec:experimental_setup} (Experimental Setup) and \cref{subsec:method} (Method) from the main article to facilitate reproducibility.
We perform training and inference on a cluster utilising $4\times$ NVIDIA A100 GPUs with 40~GB of memory each.
The training memory and runtime footprints are specified in the following subsections.

\subsection{Details on Subset Creation}
\label{sup_subsec:subset_creation}
In total, VEIS~\cite{Saleh_2018_ECCV} contains 61305 images, of which 30180 depict a multi-class scene, which is relevant for our task.
The data is captured as a single video sequence of a camera trajectory through a static scene, making consecutive frames very similar.
Thus, to reduce computational requirements of generating data, while maximising variety, we extract every 10th video frame and add it to our final subset, yielding 3018 images.

SHIFT~\cite{sun2022shift} is a collection of multiple video sequences and contains 2.5 million images in total.
We utilise rare class sampling (RCS) introduced in \cite{Hoyer_2022_CVPR} to obtain a subset of 3000 images.
In contrast to \cite{Hoyer_2022_CVPR}, we sample without replacement and use an RCS temperature $T=0.05$.
The pool of images from which the subset is chosen depends on the limitations of a given method.
For competitors that use the RGB image (CycleGAN~\cite{zhu2017unpaired}, MUNIT~\cite{huang2018multimodal}, Photorealism Enhancement~\cite{richter2022enhancing}, and I2I-Turbo~\cite{parmar2024one}), we sample only from images captured under normal daytime conditions (clear, cloudy, overcast).
Had we kept the complete image pool, these methods would also have to learn to re-adjust the adverse condition in the synthetic sample towards the target dataset (e.g., night to day adjustment when translating to Cityscapes~\cite{Cordts2016Cityscapes}).
This comes at the cost of reduced image variety.
In contrast, the remainder of the methods either only use the semantic label (Ours, EnCo~\cite{cai2024rethinking}) or do not require training (Original) and are thus not affected by the appearance of the RGB image.
Therefore, they sample from the full image pool.

\subsection{Details on the Proposed Framework}
\label{sup_subsec:method_details}
The pseudo-labeller $f_{L}$, HRDA~\cite{hoyer2022hrda}, is trained using the official codebase and hyperparameters, with a small adaptation to handle multiple source datasets.
Namely, the statistics for rare class sampling (RCS; see \cite{hoyer2022hrda}) are computed for each of the three synthetic datasets separately.
During training, a data batch is then formed as follows:
First, one of the synthetic datasets is chosen with a probability proportional to its size (i.e., GTA5~\cite{Richter_2016_ECCV} with 24966 images is more likely to be chosen than UrbanSyn~\cite{GOMEZ2025130038} with 7539 images).
Then, RCS is employed on that dataset to obtain one image for the batch.
This process then repeats.
Training of HRDA on $1\times$A100 GPU, requires 24 GB of GPU memory, and takes 20 hours.

The training of our model is composed of two stages: (i) the fine-tuning of Stable Diffusion 2.1~\cite{rombach2022high} towards the target distribution and (ii) the conditioning with pseudo labels. For the fine-tuning towards Cityscapes~\cite{Cordts2016Cityscapes}, we chose a batch size of 12 and train first for 240k training steps on a resolution of $384\times 768$ before we continue with a batch size of 8 and a resolution of $512 \times 1024$ for 150k more training steps. With $4\times$A100 GPUs, the first stage takes 50 hours + 54 hours. For the conditioning stage, we add 4 channels to the input layer of our model and concatenate the guiding condition in latent space to the noisy input. We keep the training resolution unchanged and train for 120k steps with a batch size of 8. With $4\times$A100 GPUs, the second stage is completed within 48 hours. Throughout training, we use a constant learning rate of $2\times 10^{-5}$.

For the transfer towards ACDC~\cite{sakaridis2021acdc} as the target data, we fine-tune our final Cityscapes model for 6k steps with a resolution of $512\times 1024$ and a batch size of 8, as in the previous conditioning stage. Because of the limited number of 1600 ACDC images, further training would result in overfitting towards the training samples. On $4\times$A100 GPUs, the transfer takes only 2.5 hours.

During training, our model is trained with width to height ratios of 2:1. However, the aspect ratios of unseen arbitrary synthetic datasets can be different. During inference, before generating an image with a synthetic semantic map, we extract the largest possible centre crop with the width to height ratio of 2:1 to guarantee a consistent generative quality by not risking out-of-distribution output sizes. Meanwhile, competing approaches translate the whole image, without potentially ignoring border regions.

\subsection{Details on the Downstream Task}
\label{sup_subsec:downstream_details}
{\sloppy We use the existing implementations of SegFormer/MiT-B5~\cite{NEURIPS2021_64f1f27b} and DeepLabV3+/R101-D8~\cite{chen2018encoder} available in the mmsegmentation v1.2.2~\cite{mmseg2020} framework.
During training, we utilise the exact data augmentation pipeline used in SegFormer and set the crop size to 1024$\times$1024 for both target datasets.
In contrast to \cite{NEURIPS2021_64f1f27b}, a batch size of 2 is used to make the experiments feasible with the given resources in a reasonable timeframe.
Note that just the final comparison table (\cref{tab:sota-comparison-downstream}) consists of 98 trained models, while the full research project included additional trained models that did not make it into the paper for brevity reasons.
Evaluation is performed at the original image resolution without sliding windows.
We use an AdamW~\cite{loshchilov2018decoupled} optimiser with 160K and 40K optimisation steps for SegFormer and DeepLabV3+, respectively.
Linear learning rate warmup from $6\times10^{-6}$ to $6\times10^{-5}$ is used in the first 1500 steps, followed by a linear decay to zero in the remaining steps.
On $1\times$A100 GPU, SegFormer requires 21.5 hours and 26 GB of GPU memory for a single training run, while DeepLabV3+ requires 4.5 hours and 13 GB.
Inference on the test set takes only a few minutes.
\par}

\subsection{Details on the FID and CMMD Operating Resolution}
The FID~\cite{heusel2017gans} an CMMD~\cite{jayasumana2024rethinking} scores are computed at the resolution of the generated data, i.e., original image resolution of the synthetic image for all the competitors, and $512\times1024$ resolution for our model.

\section{Details on Comparison to DGInStyle and Instance Augmentation}
\label{sup_sec:dginstyle_and_instaug}
This section contains details on the experiment from \cref{subsec:quantitative_and_qualitative_comparison}, where we compare to DGInStyle~\cite{jia2024dginstyle} and Instance Augmentation~\cite{kupyn2024dataset}.
DGInStyle and Instance Augmentation are diffusion-based representatives of the subfields of domain generalisation and data augmentation research, respectively.
Since neither method is designed for domain adaptation, we observe, as expected, that our approach clearly outperforms them. Both methods are computationally expensive since they run their generation pipeline multiple times on a single image.
Thus, for DGInStyle we opted to use only 7000 images translated from GTA5~\cite{Richter_2016_ECCV} provided by the authors and train SegFormer~\cite{NEURIPS2021_64f1f27b}, while monitoring that no over-fitting occurs. Compared to our model with GTA5 as source (see \cref{tab:sota-comparison-downstream}), we observe a -3.8\%pt. mIoU drop for Cityscapes~\cite{Cordts2016Cityscapes} as the target domain. Since DGInStyle data also includes adverse scenarios, we additionally compare it to our model with UrbanSyn~\cite{GOMEZ2025130038} as source and ACDC~\cite{sakaridis2021acdc} as the target domain; we observe a drop of -4.6\%pt. mIoU compared to our approach.
For Instance Augmentation, we run their official pipeline on UrbanSyn and observe a -6.2\%pt. mIoU drop compared to our approach using SegFormer and Cityscapes as a target.

\section{Additional Quantitative Results to the Main Comparison Table}
\label{sup_subsec:additional_quantitative}
We add details to the main comparison table (\cref{tab:sota-comparison-downstream}) from \cref{subsec:quantitative_and_qualitative_comparison}.

As stated in \cref{subsec:experimental_setup}, all translation methods perform worse when paired with original semantic labels instead of pseudo-labels in the downstream task (see \cref{tab:original-vs-pseudo-label-results}).

\begin{table}
    \centering
    \setlength{\tabcolsep}{1mm}
    \caption{\emph{Comparison between using original synthetic semantic labels versus pseudo-labels on semantic segmentation performance} (mIoU in \%, $\uparrow$). Our method is compared to five competing image translation methods, translating from five synthetic datasets to Cityscapes~\cite{Cordts2016Cityscapes}. All methods perform better when using pseudo-labels.}
    \begin{tabular}{@{}clccccc@{}}
    \toprule
    Label & Translation & \multicolumn{5}{c}{$\rightarrow$ Cityscapes (mIoU in \%, $\uparrow$)} \\
    \cmidrule(lr){3-7}
    Type & Method & VEIS & SHIFT & GTA5 & Synscapes & UrbanSyn \\
    \midrule
    \multirow{8}{*}{\rotatebox[origin=c]{90}{Original}}
        & CycleGAN
            & 44.9 
            & 45.8 
            & 53.0 
            & 56.2 
            & 66.2 
            \\
        & MUNIT      
            & 45.6 
            & 45.6 
            & 50.7 
            & 55.9 
            & 64.6 
            \\
        & Ph. Enhanc. 
            & 45.6 
            & 47.6 
            & 52.5 
            & 56.8 
            & 66.2 
            \\
        & I2I-Turbo   
            & 45.7 
            & 46.2 
            & 50.0 
            & 55.1 
            & 65.2 
            \\
        & EnCo     
            & 15.7 
            & 16.4 
            & 18.9 
            & 13.0 
            & 17.3 
            \\
        \cmidrule(lr){2-7}
        & Ours   
            & 59.2 
            & 51.3 
            & 52.6 
            & 60.2 
            & 66.4 
            \\    
        \midrule
        \multirow{8}{*}{\rotatebox[origin=c]{90}{Pseudo}}
        & CycleGAN
            & 65.6 
            & 63.8 
            & 67.9 
            & 70.3 
            & 71.0 
            \\
        & MUNIT   
            & 64.1 
            & 62.9 
            & 65.9 
            & 65.3 
            & 71.2 
            \\
        & Ph. Enhanc.
            & 62.4 
            & 61.3 
            & 64.6 
            & 68.4 
            & 71.0 
            \\
        & I2I-Turbo
            & 60.0 
            & 61.5 
            & 63.4 
            & 64.4 
            & 69.6 
            \\
        & EnCo      
            & 34.3 
            & 34.9 
            & 32.9 
            & 33.2 
            & 29.1 
            \\
        \cmidrule(lr){2-7}
        & Ours   
            & 72.8 
            & 69.0 
            & 70.1 
            & 73.8 
            & 72.9 
            \\
    \bottomrule
    \end{tabular}
    \label{tab:original-vs-pseudo-label-results}
\end{table}

Per-class results of the compared methods are provided separately for each target dataset and the downstream model: SegFormer~\cite{NEURIPS2021_64f1f27b}, translating from five synthetic datasets to Cityscapes~\cite{Cordts2016Cityscapes} in \cref{tab:segformer-cityscapes-classwise}, and to ACDC~\cite{sakaridis2021acdc} in \cref{tab:segformer-acdc-classwise}; DeepLabV3+~\cite{chen2018encoder} results are shown in \cref{tab:deeplabv3plus-cityscapes-classwise} and \cref{tab:deeplabv3plus-acdc-classwise}, respectively.

Note that VEIS~\cite{Saleh_2018_ECCV} and SHIFT~\cite{sun2022shift} do not contain all classes present in the target datasets, thus, we compute the mean intersection over union (mIoU) only over the existing ones.
The accuracy of the missing classes is denoted with ``--''.
Our method reaches approximately the same accuracy over all five synthetic datasets for most classes.
There are, however, a few exceptions.
For example, compared to other source datasets, a larger performance drop is observed when translating from SHIFT to Cityscapes for the ``Trck'' (\cref{tab:deeplabv3plus-cityscapes-classwise}) and ``Bus'' (\cref{tab:segformer-cityscapes-classwise}, \cref{tab:deeplabv3plus-cityscapes-classwise}) classes.
We attribute this to the fact that the types of trucks and buses do not structurally match the target distribution.
SHIFT contains only minibuses and light trucks, while in Cityscapes, buses are usually larger, and there are also heavy trucks, both of which exhibit a completely different shape.
A similar limitation exists in GTA5~\cite{Richter_2016_ECCV}, where the ``Trck'' class is biased toward American-style trucks. The performance drop in the ``Train'' class can be explained by its rarity in GTA5.
We conclude that achieving favourable performance requires asset shapes that closely match the target distribution.
Additionally, care should be taken to ensure that rare classes are well represented in the synthetic data.
Both aspects would fall within the responsibility of the 3D modelling community in the proposed collaborative framework.

\begin{table*}
    \footnotesize
    \caption{\emph{Comparison on per-class semantic segmentation performance} (mIoU in \%) \emph{using SegFormer}~\cite{NEURIPS2021_64f1f27b}.
    Our method is compared to five competing image translation methods, translating from five synthetic datasets to Cityscapes~\cite{Cordts2016Cityscapes}. Methods marked with \dag~can generate multiple diverse images per synthetic sample. For these, we generate three images per sample. All other methods are deterministic and produce only one image per sample. The best approach is highlighted in \textbf{bold}, the second best \underline{underlined}.}
    \centering
    \footnotesize
    \setlength{\tabcolsep}{1mm}
    \resizebox{\textwidth}{!}{
    \begin{tabular}{@{}l|c|ccccccccccccccccccc@{}}
    \toprule
    Method & mIoU & Rd.\! & Sdwk & Bldg & Wall & Fnc & Pole & TLgt & TSign & Veg & Terr &
Sky & Pers & Rdr & Car & Trck & Bus & Train & Mcy & Bike \\
    \midrule
    \multicolumn{21}{c}{VEIS $\rightarrow$ Cityscapes} \\
    \midrule
    Original & 34.5 & 71.4 & 13.3 & 61.4 & - & - & 9.9 & 28.1 & 38.5 & 74.9 & 19.3 & 75.5 & 53.5 & 14.2 & 51.8 & 22.9 & 22.3 & 0.5 & 10.1 & 18.1 \\
    CycleGAN & 65.6 & 93.8 & 56.6 & 88.4 & - & - & 51.9 & \underline{61.7} & 65.3 & \underline{90.7} & 45.8 & 93.5 & \underline{75.9} & 43.6 & \underline{92.3} & 65.3 & 58.2 & 7.0 & \underline{52.8} & \underline{72.9} \\
    MUNIT\textsuperscript{\dag} & \underline{66.4} & \underline{94.4} & \underline{61.6} & \underline{89.2} & - & - & \underline{53.9} & \textbf{63.9} & \underline{68.8} & \textbf{90.8} & \underline{48.5} & \underline{94.1} & 75.6 & 41.8 & 91.7 & \underline{66.0} & \underline{62.6} & 10.5 & 44.3 & 70.4 \\
    Ph. Enhanc. & 62.4 & 90.7 & 38.9 & 86.6 & - & - & 47.4 & 57.1 & 66.4 & 89.0 & 35.0 & 92.6 & 73.4 & \underline{44.1} & 90.5 & 64.4 & 58.8 & 12.8 & 44.3 & 69.7 \\
    I2I-Turbo & 60.0 & 84.2 & 42.8 & 75.5 & - & - & 49.0 & 59.5 & 65.8 & 88.5 & 37.3 & 88.9 & 74.2 & 39.9 & 89.7 & 50.1 & 51.2 & \underline{14.0} & 40.8 & 68.8 \\
    EnCo & 34.3 & 92.9 & 53.7 & 72.7 & - & - & 11.0 & 0.0 & 0.0 & 84.8 & 47.0 & 93.1 & 0.0 & 0.0 & 79.3 & 19.5 & 13.3 & 0.3 & 0.1 & 14.8 \\
    Ours\textsuperscript{\dag} & \textbf{74.4} & \textbf{96.3} & \textbf{72.3} & \textbf{90.2} & - & - & \textbf{54.7} & 61.5 & \textbf{70.2} & 90.6 & \textbf{50.2} & \textbf{94.5} & \textbf{78.6} & \textbf{54.1} & \textbf{93.2} & \textbf{69.4} & \textbf{80.7} & \textbf{76.4} & \textbf{57.6} & \textbf{74.3} \\
    \midrule
    \multicolumn{21}{c}{SHIFT $\rightarrow$ Cityscapes} \\
    \midrule
    Original & 47.8 & 94.4 & 63.2 & 83.3 & 9.6 & 2.5 & 46.7 & 35.2 & 44.6 & 85.4 & 23.9 & 87.4 & 69.1 & 37.8 & 87.2 & 28.7 & 2.3 & - & 28.3 & 31.1 \\
    CycleGAN & 63.8 & \underline{96.4} & 73.6 & \underline{89.1} & \underline{44.5} & 41.4 & \underline{55.8} & \underline{59.4} & 50.9 & \underline{91.0} & \underline{51.1} & \underline{94.3} & \underline{74.8} & 37.9 & 90.9 & 43.3 & 42.3 & - & 44.5 & \underline{68.0} \\
    MUNIT\textsuperscript{\dag} & \underline{64.9} & 96.3 & \underline{73.9} & 88.7 & 43.3 & \underline{43.3} & 54.2 & 54.8 & 49.7 & 90.6 & 50.9 & 94.1 & 73.6 & 41.4 & \underline{91.7} & \underline{52.3} & \underline{54.3} & - & \underline{48.5} & 65.6 \\
    Ph. Enhanc. & 61.3 & 95.4 & 69.2 & 88.1 & 32.9 & 38.7 & 54.3 & 50.1 & 50.7 & 90.5 & 48.8 & 93.3 & 73.5 & \underline{42.8} & 90.3 & 42.0 & 35.7 & - & 43.0 & 64.3 \\
    I2I-Turbo & 61.5 & 95.3 & 68.3 & 88.2 & 40.7 & 38.8 & 54.9 & 56.0 & \underline{51.9} & 90.6 & 50.5 & 93.4 & 72.2 & 40.7 & 90.1 & 36.1 & 34.0 & - & 39.5 & 65.4 \\
    EnCo & 34.9 & 93.1 & 57.3 & 71.4 & 22.0 & 25.6 & 7.0 & 0.0 & 2.7 & 80.4 & 47.0 & 91.7 & 1.0 & 0.0 & 80.9 & 27.1 & 20.7 & - & 0.3 & 0.0 \\
    Ours\textsuperscript{\dag} & \textbf{70.6} & \textbf{96.5} & \textbf{74.6} & \textbf{90.7} & \textbf{54.2} & \textbf{51.1} & \textbf{57.4} & \textbf{61.0} & \textbf{64.7} & \textbf{91.4} & \textbf{53.4} & \textbf{94.6} & \textbf{77.8} & \textbf{49.3} & \textbf{93.3} & \textbf{68.5} & \textbf{70.2} & - & \textbf{51.6} & \textbf{71.5} \\
    \midrule
    \multicolumn{21}{c}{GTA5 $\rightarrow$ Cityscapes} \\
    \midrule
    Original & 47.2 & 76.5 & 25.0 & 83.0 & 29.6 & 34.0 & 32.7 & 52.4 & 23.7 & 86.7 & 39.5 & 87.8 & 70.3 & 33.3 & 86.1 & 31.0 & 37.6 & 3.0 & 32.1 & 33.3 \\
    CycleGAN & \underline{67.9} & \underline{96.7} & \underline{75.2} & \underline{90.8} & 54.8 & \underline{52.7} & \textbf{57.6} & \textbf{62.7} & 59.7 & \textbf{91.4} & 53.0 & 94.6 & \textbf{76.3} & 42.7 & \underline{92.2} & 60.0 & \underline{66.7} & \underline{39.6} & \underline{52.1} & \underline{70.8} \\
    MUNIT\textsuperscript{\dag} & 65.8 & 96.4 & 73.8 & 90.7 & \textbf{58.1} & 52.6 & 56.3 & \underline{62.5} & \underline{62.1} & \underline{91.2} & \underline{53.4} & 94.4 & 75.3 & 44.2 & 91.7 & 55.7 & 65.4 & 17.8 & 42.8 & 65.9 \\
    Ph. Enhanc. & 64.6 & 93.9 & 63.7 & 90.4 & \underline{55.5} & 51.0 & 56.6 & 59.4 & 56.4 & 91.0 & 51.6 & \underline{94.7} & 74.7 & 39.6 & 91.7 & \underline{62.4} & 63.9 & 26.5 & 45.1 & 59.2 \\
    I2I-Turbo & 63.4 & 94.3 & 64.6 & 90.3 & 51.9 & 49.4 & 54.6 & 61.8 & 59.7 & 90.9 & 51.6 & 94.1 & 75.8 & \underline{44.8} & 91.3 & 51.5 & 58.4 & 16.9 & 37.9 & 64.1 \\
    EnCo & 32.9 & 93.7 & 61.9 & 67.2 & 26.1 & 33.1 & 4.6 & 0.0 & 2.3 & 69.8 & 49.3 & 91.2 & 0.9 & 0.0 & 79.5 & 34.2 & 11.0 & 0.0 & 0.0 & 0.0 \\
    Ours\textsuperscript{\dag} & \textbf{70.8} & \textbf{97.0} & \textbf{77.4} & \textbf{90.9} & 52.8 & \textbf{54.1} & \underline{57.5} & 60.3 & \textbf{67.1} & \textbf{91.4} & \textbf{54.0} & \textbf{94.8} & \underline{76.2} & \textbf{46.2} & \textbf{92.5} & \textbf{63.2} & \textbf{80.3} & \textbf{62.1} & \textbf{55.9} & \textbf{72.1} \\
    \midrule
    \multicolumn{21}{c}{Synscapes $\rightarrow$ Cityscapes} \\
    \midrule
    Original & 55.0 & 92.5 & 50.7 & 81.8 & 33.5 & 38.1 & 51.4 & 54.7 & 60.9 & 88.0 & 40.1 & 89.4 & 71.4 & 36.3 & 90.3 & 23.1 & 19.8 & 19.9 & 39.5 & 64.6 \\
    CycleGAN & \underline{70.3} & \textbf{96.8} & \underline{75.9} & \underline{89.9} & \underline{50.0} & \underline{45.7} & 56.7 & 62.7 & 64.8 & \textbf{91.2} & \underline{51.2} & \textbf{94.0} & 77.8 & 48.1 & 91.4 & 59.2 & \underline{73.4} & \underline{75.7} & \underline{57.8} & 73.2 \\
    MUNIT\textsuperscript{\dag} & 67.3 & \underline{95.5} & 67.5 & 88.7 & 45.1 & 37.8 & 57.2 & 63.3 & 66.7 & 90.7 & 50.2 & \underline{93.7} & \underline{78.4} & 49.1 & 91.8 & 46.6 & 52.5 & 72.9 & 56.4 & \underline{74.8} \\
    Ph. Enhanc. & 68.4 & 94.4 & 65.0 & 89.0 & 48.3 & 39.2 & \underline{57.6} & \textbf{64.1} & \textbf{69.0} & \underline{90.9} & 49.1 & 93.4 & 77.7 & \underline{51.1} & \underline{92.1} & \underline{64.7} & 61.7 & 62.7 & 56.8 & 73.2 \\
    I2I-Turbo & 64.4 & 94.9 & 65.6 & 87.8 & 40.9 & 36.9 & 56.4 & 63.8 & 66.3 & 90.6 & 49.2 & 92.3 & 78.1 & \underline{51.1} & 91.8 & 41.0 & 43.8 & 44.2 & 54.3 & 74.4 \\
    EnCo & 33.2 & 94.3 & 64.9 & 69.3 & 28.1 & 31.0 & 3.1 & 0.0 & 0.0 & 77.7 & 50.4 & 92.3 & 0.0 & 0.0 & 78.2 & 29.0 & 12.3 & 0.0 & 0.0 & 0.0 \\
    Ours\textsuperscript{\dag} & \textbf{73.6} & \textbf{96.8} & \textbf{77.2} & \textbf{90.9} & \textbf{53.3} & \textbf{50.6} & \textbf{58.7} & \underline{63.9} & \underline{68.9} & \textbf{91.2} & \textbf{53.7} & 92.9 & \textbf{79.7} & \textbf{54.5} & \textbf{93.3} & \textbf{69.2} & \textbf{83.6} & \textbf{81.5} & \textbf{62.6} & \textbf{75.3} \\
    \midrule
    \multicolumn{21}{c}{UrbanSyn $\rightarrow$ Cityscapes} \\
    \midrule
    Original & 61.7 & 91.2 & 49.6 & 87.2 & 21.7 & 45.4 & 53.9 & 61.4 & 69.1 & 87.2 & 32.7 & 89.6 & 76.3 & 52.3 & 92.2 & 70.1 & 59.6 & 21.7 & 39.1 & 71.5 \\
    CycleGAN & \underline{71.0} & \underline{96.9} & \underline{76.5} & \underline{90.7} & \textbf{47.0} & 51.2 & 57.4 & 63.9 & 68.0 & \textbf{91.4} & \textbf{52.5} & \underline{94.3} & 78.6 & 50.8 & 93.1 & 69.3 & 78.1 & 57.8 & 58.2 & 73.1 \\
    MUNIT\textsuperscript{\dag} & 70.9 & 95.8 & 69.6 & 90.6 & \underline{45.6} & \underline{52.5} & \underline{59.5} & \underline{65.1} & \textbf{72.4} & 91.1 & 50.6 & 94.0 & \textbf{80.1} & \underline{55.0} & \textbf{93.7} & \textbf{74.7} & 75.2 & 46.6 & \textbf{61.4} & 74.2 \\
    Ph. Enhanc. & \underline{71.0} & 94.5 & 64.8 & 90.2 & 40.3 & 51.0 & 58.9 & 65.0 & \underline{71.0} & 90.9 & \underline{51.9} & 93.5 & 79.1 & \textbf{55.5} & \underline{93.6} & \underline{74.5} & \underline{78.4} & \underline{61.6} & 60.1 & 73.9 \\
    I2I-Turbo & 69.6 & 94.8 & 65.9 & 90.4 & 45.2 & 48.8 & \textbf{59.8} & \textbf{65.4} & 70.4 & \underline{91.2} & 47.1 & 92.7 & 78.1 & 51.9 & \underline{93.6} & 72.6 & 74.6 & 46.7 & 57.7 & \textbf{74.9} \\
    EnCo & 29.1 & 91.4 & 53.7 & 66.3 & 14.0 & 19.0 & 0.5 & 0.0 & 0.0 & 66.2 & 45.5 & 89.8 & 2.0 & 0.0 & 79.5 & 19.5 & 2.5 & 0.0 & 2.6 & 0.0 \\
    Ours\textsuperscript{\dag} & \textbf{73.0} & \textbf{97.1} & \textbf{78.3} & \textbf{90.9} & 45.3 & \textbf{54.5} & 58.5 & 63.7 & 70.5 & 90.3 & 51.1 & \textbf{94.4} & \underline{79.4} & 54.3 & 93.2 & 66.3 & \textbf{84.8} & \textbf{78.5} & \underline{61.0} & \underline{74.5} \\
    \bottomrule
    \end{tabular}
    }
    \label{tab:segformer-cityscapes-classwise}
\end{table*}

\begin{table*}
    \footnotesize
    \caption{\emph{Comparison on per-class semantic segmentation performance} (mIoU in \%) \emph{using SegFormer}~\cite{NEURIPS2021_64f1f27b}.
    Our method is compared to five competing image translation methods, translating from two synthetic datasets to ACDC~\cite{sakaridis2021acdc}. The best approach is highlighted in \textbf{bold}, the second best \underline{underlined}.}
    \centering
    \footnotesize
    \setlength{\tabcolsep}{1mm}
    \resizebox{\textwidth}{!}{
    \begin{tabular}{@{}l|c|ccccccccccccccccccc@{}}
    \toprule
    Method & mIoU & Rd.\! & Sdwk & Bldg & Wall & Fnc & Pole & TLgt & TSign & Veg & Terr &
Sky & Pers & Rdr & Car & Trck & Bus & Train & Mcy & Bike \\
    \midrule
    \multicolumn{21}{c}{VEIS $\rightarrow$ ACDC} \\
    \midrule
    Original & 18.5 & 9.2 & \underline{16.2} & 30.8 & - & - & 6.5 & 37.9 & 24.1 & 43.4 & 18.2 & 58.9 & 23.2 & 0.9 & 34.0 & 4.2 & 3.7 & 0.3 & 3.7 & 0.2 \\
    CycleGAN & 44.7 & \underline{71.3} & 4.1 & \textbf{77.0} & - & - & \textbf{47.2} & 36.5 & \textbf{50.3} & \underline{68.7} & \underline{32.1} & 81.3 & 50.9 & 11.9 & 80.4 & 59.1 & 39.2 & 5.3 & 25.7 & 19.2 \\
    MUNIT & 46.4 & 70.0 & 3.0 & \underline{75.9} & - & - & 40.5 & \underline{39.9} & \underline{49.4} & 67.9 & 30.9 & 81.5 & \underline{51.4} & \underline{21.9} & \textbf{81.3} & 61.5 & 43.9 & 22.9 & \underline{26.7} & \underline{19.5} \\
    Ph. Enhanc. & \underline{48.5} & 70.1 & 0.5 & 74.1 & - & - & 42.0 & 34.8 & 46.5 & 68.2 & \textbf{33.3} & 81.2 & \textbf{53.3} & 18.2 & \underline{81.0} & \underline{65.8} & \underline{56.7} & \underline{37.3} & \textbf{36.6} & \textbf{24.6} \\
    I2I-Turbo & 43.0 & 70.3 & 0.3 & 72.6 & - & - & 36.0 & 39.4 & 47.7 & 68.5 & 31.9 & \underline{81.8} & 49.0 & 19.1 & 78.8 & 50.6 & 40.3 & 3.6 & 24.4 & 16.9 \\
    EnCo & 28.2 & 69.8 & 0.8 & 68.9 & - & - & \underline{44.3} & 22.1 & 29.9 & 63.6 & 29.7 & 79.8 & 0.0 & 0.0 & 56.3 & 8.9 & 4.4 & 0.8 & 0.0 & 0.0 \\
    Ours & \textbf{50.3} & \textbf{75.5} & \textbf{19.3} & 57.8 & - & - & 32.9 & \textbf{62.6} & 43.3 & \textbf{69.7} & 26.4 & \textbf{81.9} & 45.8 & \textbf{22.1} & \textbf{81.3} & \textbf{67.9} & \textbf{73.1} & \textbf{53.4} & 26.4 & 16.1 \\
    \midrule
    \multicolumn{21}{c}{UrbanSyn $\rightarrow$ ACDC} \\
    \midrule
    Original & 34.0 & 66.1 & \textbf{26.3} & 44.3 & 7.9 & 12.3 & 24.6 & \textbf{59.3} & 44.2 & 43.1 & 16.4 & 64.5 & 34.8 & 22.2 & 53.2 & 59.0 & 25.3 & 9.8 & 22.2 & 11.1 \\
    CycleGAN & \underline{49.8} & \textbf{72.4} & \underline{4.3} & \textbf{77.5} & \textbf{36.5} & 22.2 & \underline{52.2} & 36.3 & \underline{52.2} & 68.7 & \underline{32.4} & \underline{81.4} & 57.8 & 27.2 & \textbf{84.1} & 63.7 & \underline{65.9} & 48.5 & 35.2 & \underline{28.4} \\
    MUNIT & 49.0 & \underline{71.7} & 3.9 & 76.3 & 30.1 & \textbf{23.9} & 51.6 & \underline{40.1} & \textbf{54.0} & 68.2 & 32.0 & \textbf{81.5} & \underline{58.0} & 28.7 & \underline{83.6} & \underline{69.9} & 59.0 & 30.9 & \textbf{40.2} & 27.6 \\
    Ph. Enhanc. & 47.5 & 70.3 & 2.1 & \underline{77.3} & 27.0 & 21.4 & \textbf{52.8} & 37.6 & 49.8 & \textbf{69.4} & 32.3 & 81.2 & \textbf{58.7} & \underline{29.5} & 81.9 & 68.9 & 53.4 & 19.3 & \underline{39.0} & \textbf{30.3} \\
    I2I-Turbo & 48.1 & 70.7 & 3.1 & 76.7 & \underline{31.7} & 22.0 & 51.1 & 37.3 & 52.1 & \underline{69.0} & 31.2 & 81.3 & 57.7 & 28.9 & 82.2 & 66.6 & 52.7 & 37.6 & 37.1 & 24.9 \\
    EnCo & 28.0 & 70.0 & 1.9 & 67.4 & 20.3 & 3.9 & 39.7 & 20.6 & 28.9 & 56.7 & 26.7 & 76.8 & 0.3 & 0.0 & 57.6 & 7.8 & 1.9 & \underline{50.9} & 0.0 & 0.0 \\
    Ours & \textbf{50.4} & \textbf{72.4} & 3.9 & 75.5 & 30.7 & \underline{23.2} & 49.8 & 37.9 & 51.8 & 68.3 & \textbf{34.7} & \underline{81.4} & 56.3 & \textbf{31.8} & 81.9 & \textbf{70.0} & \textbf{83.4} & \textbf{53.4} & 30.6 & 20.2 \\
    \bottomrule
    \end{tabular}
    }
    \label{tab:segformer-acdc-classwise}
\end{table*}

\begin{table*}
    \footnotesize
    \caption{\emph{Comparison on per-class semantic segmentation performance} (mIoU in \%) \emph{using DeepLabV3+}~\cite{chen2018encoder}.
    Our method is compared to five competing image translation methods, translating from five synthetic datasets to Cityscapes~\cite{Cordts2016Cityscapes}. Methods marked with \dag~can generate multiple diverse images per synthetic sample. For these, we generate three images per sample. All other methods are deterministic and produce only one image per sample. The best approach is highlighted in \textbf{bold}, the second best \underline{underlined}.}
    \centering
    \footnotesize
    \setlength{\tabcolsep}{1mm}
    \resizebox{\textwidth}{!}{
    \begin{tabular}{@{}l|c|ccccccccccccccccccc@{}}
    \toprule
    Method & mIoU & Rd.\! & Sdwk & Bldg & Wall & Fnc & Pole & TLgt & TSign & Veg & Terr &
Sky & Pers & Rdr & Car & Trck & Bus & Train & Mcy & Bike \\
    \midrule
    \multicolumn{21}{c}{VEIS $\rightarrow$ Cityscapes} \\
    \midrule
    Original & 19.0 & 6.0 & 1.6 & 33.7 & - & - & 8.3 & 27.7 & 45.5 & 69.3 & 5.6 & 21.4 & 30.9 & 5.8 & 38.6 & 8.3 & 14.0 & 0.0 & 0.6 & 6.1 \\
    CycleGAN & \underline{57.8} & \underline{91.3} & \underline{48.3} & \underline{85.3} & - & - & 44.8 & \underline{52.5} & 59.6 & \textbf{89.5} & \textbf{40.2} & \underline{91.4} & 69.1 & 35.3 & \underline{89.5} & \underline{37.1} & 50.9 & 2.4 & \underline{29.2} & \textbf{66.5} \\
    MUNIT\textsuperscript{\dag} & 56.0 & 90.0 & 40.1 & \textbf{85.5} & - & - & 43.3 & \textbf{53.7} & \underline{60.5} & 86.6 & 34.5 & \textbf{92.6} & \textbf{71.0} & \underline{40.3} & 84.4 & 30.1 & \underline{52.3} & 2.1 & 21.6 & 63.4 \\
    Ph. Enhanc. & 46.7 & 79.6 & 32.4 & 62.6 & - & - & 32.4 & 37.8 & 49.6 & 72.1 & 27.3 & 74.7 & 57.9 & 30.9 & 82.1 & 35.2 & 39.5 & \underline{4.9} & 14.1 & 60.1 \\
    I2I-Turbo & 50.1 & 78.4 & 39.3 & 76.7 & - & - & \underline{46.9} & 48.4 & 54.8 & 86.2 & 31.9 & 80.2 & 68.0 & 34.4 & 81.6 & 23.9 & 28.4 & 4.3 & 8.0 & 60.2 \\
    EnCo & 29.0 & 90.6 & 41.7 & 68.8 & - & - & 8.5 & 0.0 & 0.0 & 79.7 & 33.6 & 90.5 & 0.0 & 0.0 & 68.1 & 8.1 & 2.5 & 0.0 & 0.0 & 0.1 \\
    Ours\textsuperscript{\dag} & \textbf{62.3} & \textbf{93.7} & \textbf{55.3} & 84.7 & - & - & \textbf{47.6} & 52.2 & \textbf{66.7} & \underline{86.8} & \underline{39.2} & 89.2 & \underline{69.2} & \textbf{41.2} & \textbf{91.1} & \textbf{53.9} & \textbf{57.6} & \textbf{26.1} & \textbf{37.9} & \underline{66.3} \\
    \midrule
    \multicolumn{21}{c}{SHIFT $\rightarrow$ Cityscapes} \\
    \midrule
    Original & 44.2 & 83.9 & 48.8 & 82.5 & 9.3 & 6.8 & 42.9 & 36.4 & \underline{44.0} & 83.2 & 8.7 & 86.1 & 65.2 & 28.5 & 84.4 & 19.6 & 2.9 & - & 23.9 & 39.2 \\
    CycleGAN & \underline{55.3} & \textbf{95.1} & \textbf{66.6} & \underline{87.0} & 32.2 & 32.6 & \underline{51.2} & \underline{45.1} & 36.4 & \textbf{90.0} & \underline{46.9} & \textbf{92.7} & \textbf{70.3} & 30.1 & 86.0 & 22.1 & 14.0 & - & \textbf{33.6} & \textbf{63.7} \\
    MUNIT\textsuperscript{\dag} & 52.2 & 92.6 & 57.3 & 85.2 & \textbf{34.7} & 33.0 & 46.5 & 26.5 & 31.0 & 87.9 & 43.8 & 86.5 & 66.4 & 34.6 & \underline{88.6} & \underline{24.9} & 14.6 & - & \underline{29.0} & 55.7 \\
    Ph. Enhanc. & 52.7 & \underline{93.9} & 62.8 & 85.3 & 28.5 & \underline{33.6} & 48.0 & 31.4 & 34.6 & 87.9 & 46.4 & 90.4 & 67.6 & \underline{35.9} & 84.5 & 22.7 & \underline{14.8} & - & 26.0 & 55.3 \\
    I2I-Turbo & 51.1 & 91.6 & 56.3 & 85.7 & \underline{32.7} & 31.6 & 50.3 & 34.8 & 34.5 & 88.5 & 39.7 & \underline{91.4} & 67.2 & 32.9 & 88.1 & 20.1 & 3.1 & - & 21.0 & 49.6 \\
    EnCo & 30.2 & 91.6 & 46.2 & 70.7 & 1.0 & 25.3 & 4.2 & 0.0 & 2.5 & 79.2 & 40.6 & 90.7 & 0.4 & 0.0 & 77.2 & 13.6 & 0.0 & - & 0.0 & 0.0 \\
    Ours\textsuperscript{\dag} & \textbf{58.3} & \underline{93.9} & \underline{66.3} & \textbf{87.9} & 29.5 & \textbf{41.1} & \textbf{52.6} & \textbf{51.6} & \textbf{55.8} & \underline{89.7} & \textbf{47.0} & 91.2 & \underline{69.3} & \textbf{38.8} & \textbf{88.9} & \textbf{30.2} & \textbf{25.1} & - & 28.7 & \underline{62.1} \\
    \midrule
    \multicolumn{21}{c}{GTA5 $\rightarrow$ Cityscapes} \\
    \midrule
    Original & 31.6 & 62.9 & 22.4 & 71.8 & 17.8 & 20.2 & 35.2 & 40.2 & 19.1 & 82.5 & 27.5 & 33.1 & 56.5 & 5.1 & 76.0 & 14.2 & 8.1 & \underline{0.8} & 6.8 & 0.0 \\
    CycleGAN & \underline{52.4} & \underline{93.7} & \underline{59.5} & \underline{88.0} & \textbf{43.4} & \underline{42.0} & \textbf{50.7} & \underline{46.7} & 41.9 & \underline{89.7} & \underline{47.6} & \underline{92.5} & \underline{64.2} & 18.2 & \underline{88.1} & \underline{36.3} & \underline{37.7} & 0.0 & \underline{14.7} & \underline{41.5} \\
    MUNIT\textsuperscript{\dag} & 47.0 & 91.3 & 52.4 & 86.8 & 34.8 & 29.6 & 43.4 & 41.9 & \underline{48.0} & 86.4 & 40.8 & 92.2 & 60.3 & 16.6 & 87.9 & 28.8 & 34.5 & 0.4 & 9.6 & 7.5 \\
    Ph. Enhanc. & 43.7 & 88.9 & 42.2 & 85.8 & 38.4 & 33.3 & 46.1 & 30.2 & 29.8 & 87.9 & 43.0 & 89.3 & 56.9 & 16.0 & 84.0 & 26.3 & 23.2 & 0.0 & 9.0 & 0.0 \\
    I2I-Turbo & 48.1 & 88.9 & 46.1 & 87.0 & 34.0 & 35.2 & 43.2 & 44.8 & 36.3 & 88.7 & 44.6 & 89.8 & 64.1 & \textbf{25.6} & 87.5 & 27.1 & 36.1 & \textbf{1.4} & 9.5 & 24.8 \\
    EnCo & 26.6 & 91.3 & 45.8 & 64.2 & 0.0 & 20.3 & 0.6 & 0.0 & 0.0 & 65.1 & 44.4 & 89.1 & 0.0 & 0.0 & 74.2 & 9.8 & 0.0 & 0.0 & 0.0 & 0.0 \\
    Ours\textsuperscript{\dag} & \textbf{55.8} & \textbf{95.4} & \textbf{66.9} & \textbf{88.9} & \underline{39.0} & \textbf{43.8} & \underline{47.3} & \textbf{49.2} & \textbf{57.8} & \textbf{89.8} & \textbf{48.9} & \textbf{93.9} & \textbf{64.4} & \underline{19.1} & \textbf{88.4} & \textbf{42.0} & \textbf{47.9} & 0.0 & \textbf{17.0} & \textbf{60.5} \\
    \midrule
    \multicolumn{21}{c}{Synscapes $\rightarrow$ Cityscapes} \\
    \midrule
    Original & 45.3 & 63.8 & 38.9 & 75.2 & 16.6 & 17.7 & 44.4 & 53.4 & 57.1 & 84.9 & 4.8 & 85.7 & 66.8 & 24.4 & 87.4 & 16.8 & 17.0 & 7.1 & 34.5 & 63.7 \\
    CycleGAN & 55.4 & \underline{94.4} & \underline{63.9} & \textbf{86.2} & \textbf{35.2} & \underline{33.7} & 47.3 & 51.1 & 57.5 & \textbf{88.5} & 39.4 & \textbf{92.1} & 68.8 & 35.4 & 87.6 & 26.3 & 40.1 & 14.7 & 28.1 & 63.0 \\
    MUNIT\textsuperscript{\dag} & 54.4 & 92.0 & 51.8 & 81.9 & 25.1 & 24.1 & 49.3 & \underline{55.8} & \underline{60.7} & 87.8 & 33.3 & \underline{90.5} & \underline{73.2} & 45.5 & 87.4 & 21.6 & 25.9 & 17.2 & 42.9 & 68.8 \\
    Ph. Enhanc. & \underline{56.2} & 92.0 & 53.8 & \underline{85.1} & 25.9 & 27.6 & 50.4 & 51.4 & 59.5 & 87.1 & 27.7 & 87.5 & 72.2 & 45.2 & \underline{88.9} & \underline{27.9} & \underline{42.9} & \underline{27.2} & \underline{45.1} & \underline{69.7} \\
    I2I-Turbo & 53.9 & 90.5 & 48.2 & 81.9 & 20.7 & 22.1 & \underline{50.5} & 48.2 & 59.3 & \underline{88.4} & 35.4 & 87.6 & 72.2 & \underline{46.3} & 87.6 & 26.2 & 33.1 & 22.7 & 36.5 & 67.6 \\
    EnCo & 26.2 & 91.0 & 46.7 & 66.5 & 0.0 & 12.6 & 0.0 & 0.0 & 0.0 & 74.4 & \underline{41.9} & 88.9 & 0.0 & 0.0 & 75.3 & 0.0 & 0.0 & 0.0 & 0.0 & 0.0 \\
    Ours\textsuperscript{\dag} & \textbf{64.1} & \textbf{95.6} & \textbf{68.8} & \textbf{86.2} & \underline{31.7} & \textbf{36.3} & \textbf{53.5} & \textbf{57.4} & \textbf{65.4} & 88.0 & \textbf{43.2} & 77.4 & \textbf{76.0} & \textbf{47.8} & \textbf{91.6} & \textbf{55.2} & \textbf{71.0} & \textbf{52.5} & \textbf{48.3} & \textbf{72.4} \\
    \midrule
    \multicolumn{21}{c}{UrbanSyn $\rightarrow$ Cityscapes} \\
    \midrule
    Original & 47.8 & 85.5 & 40.6 & 82.7 & 16.2 & 26.5 & 44.6 & 51.7 & 59.6 & 84.1 & 7.9 & 81.2 & 61.5 & 35.9 & 81.2 & 21.1 & 30.8 & 10.8 & 31.4 & 55.1 \\
    CycleGAN & 58.7 & \underline{94.0} & \underline{61.1} & \underline{88.0} & 6.3 & 35.0 & 51.5 & 55.2 & 59.6 & \textbf{89.1} & \underline{42.2} & \textbf{93.2} & 69.0 & 36.4 & 91.0 & 46.3 & 48.3 & \underline{45.9} & 35.8 & 66.6 \\
    MUNIT\textsuperscript{\dag} & \textbf{61.8} & 92.2 & 52.3 & 87.8 & 23.4 & \textbf{44.5} & \textbf{55.4} & \textbf{56.9} & \textbf{67.1} & \underline{88.7} & 40.9 & 91.6 & \textbf{76.0} & \textbf{51.9} & \underline{91.6} & 43.2 & 58.6 & 36.9 & \textbf{45.7} & \textbf{70.2} \\
    Ph. Enhanc. & \textbf{61.8} & 91.0 & 51.2 & 86.8 & \underline{29.8} & \underline{41.8} & \underline{54.8} & 53.3 & 61.7 & 87.8 & 32.1 & 89.2 & \underline{73.3} & \underline{46.2} & 91.1 & \textbf{61.2} & \textbf{66.1} & \textbf{50.8} & \underline{38.0} & \underline{68.2} \\
    I2I-Turbo & 60.0 & 91.0 & 51.6 & 87.4 & 27.2 & 38.9 & \textbf{55.4} & 54.4 & 63.2 & 88.0 & 28.3 & 89.6 & \underline{73.3} & 44.8 & 91.5 & \underline{55.4} & \underline{64.9} & 33.6 & 34.6 & 67.4 \\
    EnCo & 23.5 & 89.3 & 34.5 & 60.3 & 0.0 & 3.8 & 0.0 & 0.0 & 0.0 & 57.5 & \textbf{43.6} & 87.0 & 0.0 & 0.0 & 69.6 & 0.0 & 0.0 & 0.0 & 0.0 & 0.0 \\
    Ours\textsuperscript{\dag} & \underline{60.8} & \textbf{95.8} & \textbf{69.6} & \textbf{88.6} & \textbf{34.8} & 38.3 & 53.7 & \underline{55.3} & \underline{66.1} & 88.0 & 41.9 & \underline{92.7} & 72.6 & 43.4 & \textbf{91.8} & 48.6 & 57.2 & 16.4 & 34.7 & 66.4 \\
    \bottomrule
    \end{tabular}
    }
    \label{tab:deeplabv3plus-cityscapes-classwise}
\end{table*}

\begin{table*}
    \footnotesize
    \caption{\emph{Comparison on per-class semantic segmentation performance} (mIoU in \%) \emph{using DeepLabV3+}~\cite{chen2018encoder}.
    Our method is compared to five competing image translation methods, translating from two synthetic datasets to ACDC~\cite{sakaridis2021acdc}. The best approach is highlighted in \textbf{bold}, the second best \underline{underlined}.}
    \centering
    \footnotesize
    \setlength{\tabcolsep}{1mm}
    \resizebox{\textwidth}{!}{
    \begin{tabular}{@{}l|c|ccccccccccccccccccc@{}}
    \toprule
    Method & mIoU & Rd.\! & Sdwk & Bldg & Wall & Fnc & Pole & TLgt & TSign & Veg & Terr &
Sky & Pers & Rdr & Car & Trck & Bus & Train & Mcy & Bike \\
    \midrule
    \multicolumn{21}{c}{VEIS $\rightarrow$ ACDC} \\
    \midrule
    Original & 10.5 & 4.9 & \textbf{10.2} & 12.2 & - & - & 4.0 & 8.7 & 18.7 & 35.6 & 2.6 & 37.9 & 3.6 & 0.1 & 36.0 & 1.1 & 2.6 & 0.0 & 0.2 & 0.1 \\
    CycleGAN & 34.0 & \underline{69.5} & 3.0 & \textbf{71.4} & - & - & \textbf{29.3} & 29.6 & \textbf{40.5} & \textbf{67.9} & \underline{28.3} & 80.7 & \underline{32.1} & 0.2 & \underline{75.1} & \underline{22.8} & 14.1 & 0.0 & 2.8 & \underline{10.0} \\
    MUNIT & \textbf{36.4} & 67.4 & \underline{6.0} & \underline{70.5} & - & - & 22.0 & \underline{34.1} & \underline{39.8} & \underline{66.9} & \textbf{28.7} & \textbf{81.3} & \textbf{35.9} & \textbf{11.1} & 69.7 & \textbf{27.4} & \underline{28.6} & \underline{3.2} & \textbf{14.4} & \textbf{11.8} \\
    Ph. Enhanc. & 30.7 & 64.5 & 0.0 & 66.4 & - & - & 20.9 & 31.9 & 38.7 & 65.6 & 22.0 & 78.8 & 24.6 & 3.9 & 61.9 & 17.1 & 13.5 & 0.3 & 5.2 & 6.7 \\
    I2I-Turbo & 29.4 & 66.5 & 1.9 & 57.6 & - & - & 18.2 & 19.3 & 35.9 & 66.4 & 19.0 & 73.0 & 23.1 & 4.7 & 70.0 & 19.5 & 7.2 & 0.0 & 8.5 & 8.5 \\
    EnCo & 23.0 & 68.0 & 0.4 & 62.1 & - & - & \underline{26.6} & 16.3 & 18.0 & 61.2 & 23.2 & 79.8 & 0.0 & 0.0 & 27.5 & 8.2 & 0.0 & 0.0 & 0.0 & 0.0 \\
    Ours & \underline{34.3} & \textbf{70.8} & 3.8 & 50.7 & - & - & 17.0 & \textbf{49.4} & 32.2 & 64.0 & 23.0 & \underline{80.9} & 30.5 & \underline{7.5} & \textbf{76.3} & 20.3 & \textbf{30.7} & \textbf{4.7} & \underline{11.5} & 9.2 \\
    \midrule
    \multicolumn{21}{c}{UrbanSyn $\rightarrow$ ACDC} \\
    \midrule
    Original & 14.4 & 39.9 & \textbf{9.3} & 25.0 & 0.1 & 6.1 & 19.2 & 36.5 & 31.6 & 41.9 & 1.5 & 21.6 & 7.7 & 8.6 & 9.2 & 2.9 & 0.8 & 0.1 & 8.7 & 2.8 \\
    CycleGAN & 32.7 & \underline{67.8} & 3.0 & \underline{70.0} & 8.3 & \textbf{20.5} & 30.5 & 24.5 & 37.0 & 65.5 & 21.6 & 80.4 & 42.5 & 7.1 & 71.9 & \underline{21.8} & 19.4 & 14.8 & 1.9 & 12.4 \\
    MUNIT & \textbf{36.1} & 65.0 & 5.8 & \textbf{70.3} & \underline{17.2} & 13.4 & \underline{37.2} & \underline{40.1} & \textbf{46.6} & 63.4 & \textbf{26.9} & \underline{81.2} & \textbf{49.2} & \textbf{12.9} & \textbf{76.0} & 19.7 & \underline{25.2} & 4.9 & \textbf{14.0} & \underline{17.7} \\
    Ph. Enhanc. & \textbf{36.1} & \textbf{68.9} & \underline{6.1} & 68.8 & 14.4 & 14.3 & 35.6 & 39.9 & \underline{46.3} & \underline{67.6} & \underline{22.8} & 79.4 & \underline{43.4} & 7.6 & 75.0 & \textbf{23.4} & \textbf{27.5} & 16.6 & \underline{10.6} & \underline{17.7} \\
    I2I-Turbo & 33.1 & 64.8 & 2.6 & 67.8 & 12.7 & 13.3 & \textbf{40.2} & 28.6 & 33.9 & 65.1 & 17.7 & 80.6 & 39.4 & 7.5 & 71.3 & 20.7 & 22.5 & 13.9 & 7.7 & \textbf{18.9} \\
    EnCo & 20.5 & 67.6 & 0.2 & 56.8 & 13.0 & 0.3 & 16.9 & 19.2 & 13.1 & 50.2 & 22.6 & 76.3 & 0.0 & 0.0 & 30.8 & 2.0 & 0.0 & \textbf{21.0} & 0.0 & 0.0 \\
    Ours & \underline{34.6} & \textbf{68.9} & 5.8 & 63.3 & \textbf{17.4} & \underline{15.2} & 33.2 & \textbf{52.7} & 42.7 & \textbf{68.1} & 21.1 & \textbf{81.4} & 30.4 & \underline{9.2} & \underline{75.7} & 15.6 & 22.3 & \underline{19.0} & 8.4 & 7.7 \\
    \bottomrule
    \end{tabular}
    }
    \label{tab:deeplabv3plus-acdc-classwise}
\end{table*}

\section{Additional Qualitative Samples}
\label{sup_subsec:additional_qualitative}

\begin{figure}[htbp]
    \centering
    \includegraphics[width=.97\linewidth]{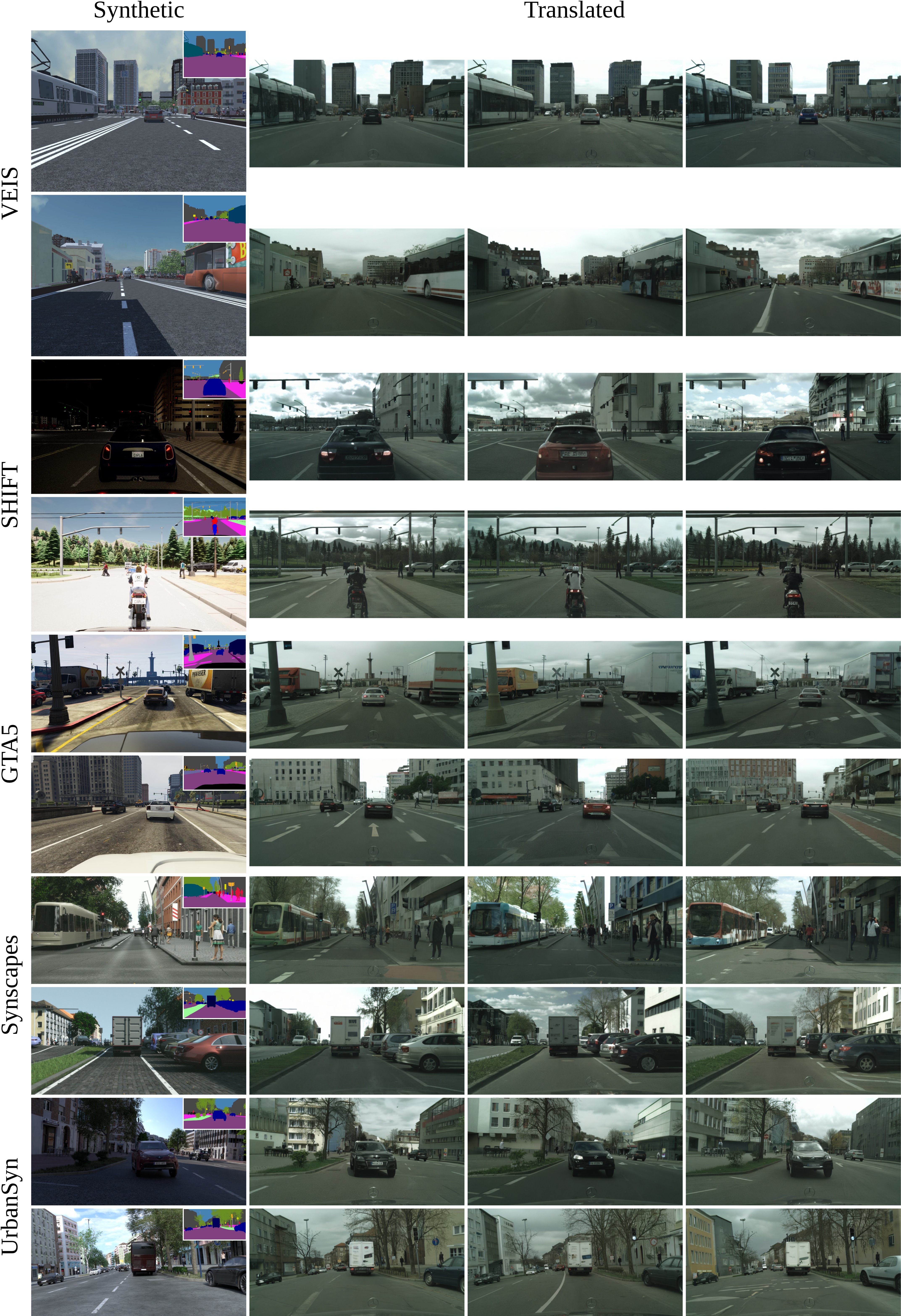}
    \caption{\emph{Sample images translated by our method from the five synthetic datasets to Cityscapes}~\cite{Cordts2016Cityscapes}. The first column contains two images per synthetic dataset and a semantic label in the top-right corner. The remaining columns show our translation for three different noise seeds.}
\label{fig:transfer_examples}
\end{figure}

{\sloppy We show additional qualitative samples of our method translating to Cityscapes~\cite{Cordts2016Cityscapes} in \cref{fig:transfer_examples} to complement \cref{fig:teaser,fig:main_qualitative} from the main paper.
As mentioned in \cref{subsec:method-filtering}, our method can translate a single semantic map into many diverse visual scenes.\par}

\begin{figure}[htbp]
    \centering
    \includegraphics[width=1\linewidth]{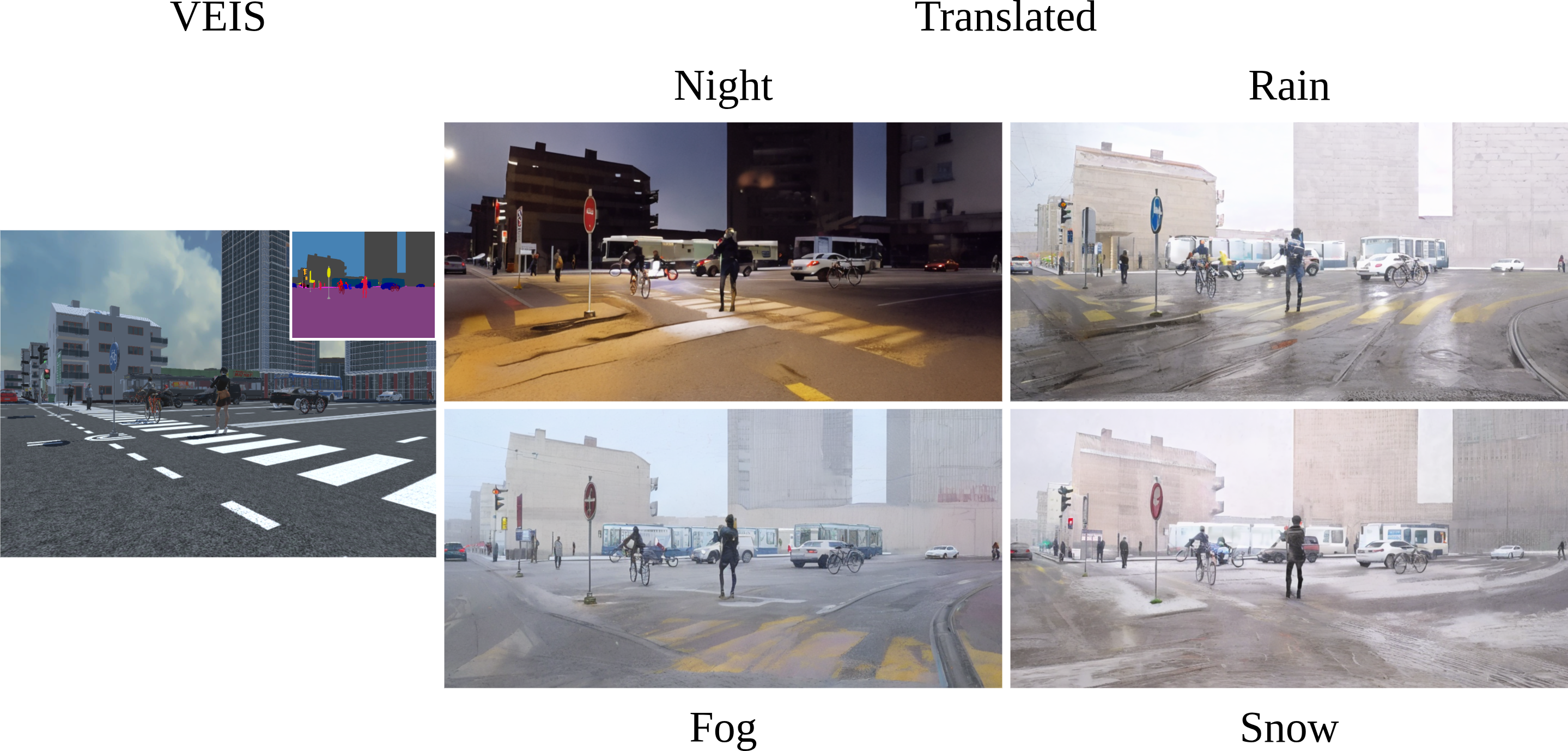}
  \hfill
    \caption{\emph{Sample images translated by our method from VEIS}~\cite{Saleh_2018_ECCV}\emph{ to ACDC}~\cite{sakaridis2021acdc}. The first column contains the synthetic image and a semantic label in the top-right corner. The remaining images show our translation for the four different weather conditions.}
\label{fig:transfer_examples_acdc}
\end{figure}

In \cref{fig:transfer_examples_acdc}, we show an example for all four weather translations of our method from VEIS~\cite{Saleh_2018_ECCV} to ACDC~\cite{sakaridis2021acdc} to complement \cref{fig:main_qualitative} from the main article.

\begin{figure}[tb]
  \centering
  \includegraphics[width=1.0\linewidth]{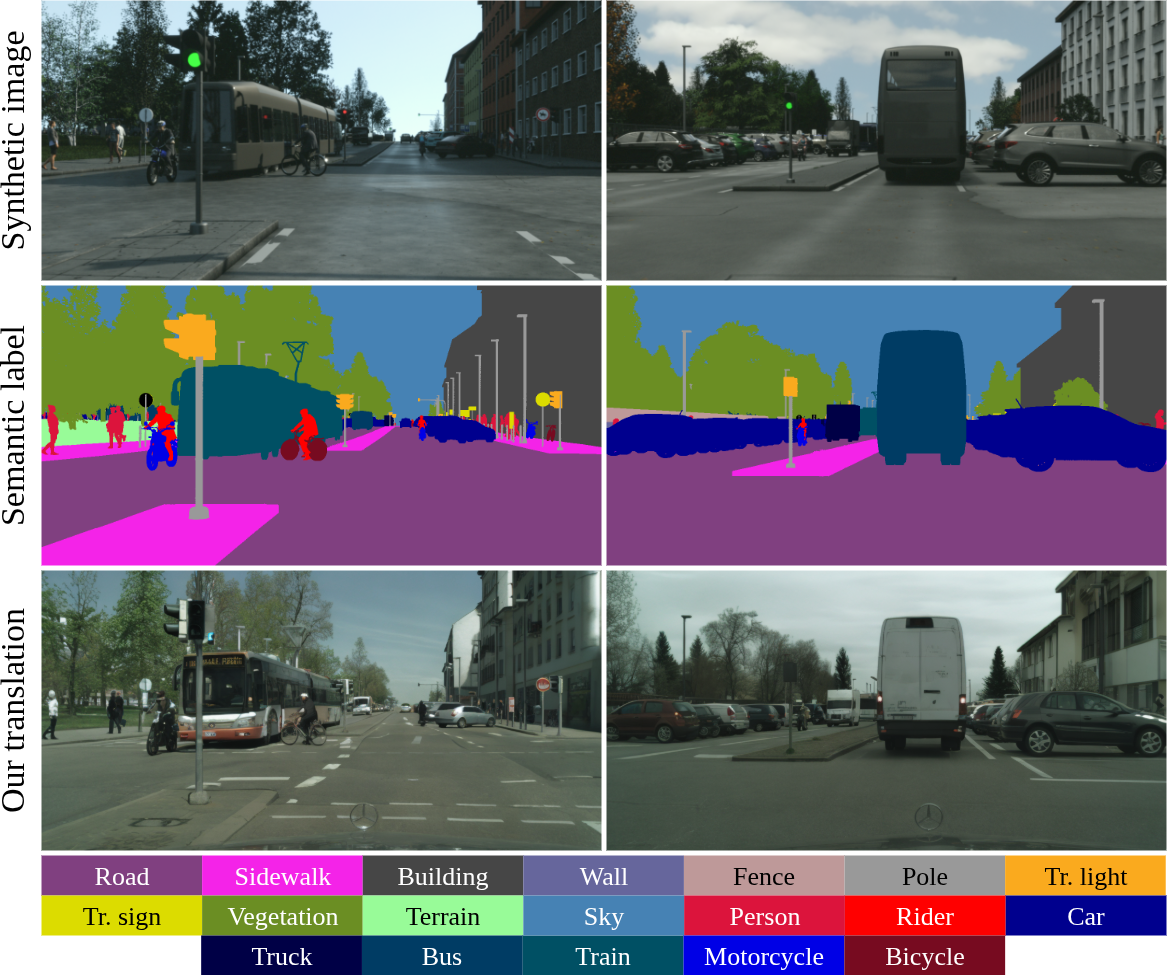}
  \caption{\emph{Two examples of a mismatch between the generation of our model and the semantic label.} A train is incorrectly generated as a ``bus'' \emph{(first column)}, and a bus is generated as a transport van belonging to the ``car'' class \emph{(second column)}. If the label is not rectified, this can negatively affect the performance of the downstream model.}
  \label{fig:failure_case}
\end{figure}

Similarly, as mentioned in the ablation study (\cref{subsec:ablation}), we illustrate the drawback of the mismatch between the generated image and the semantic label in \cref{fig:failure_case}.

\end{document}